\documentclass[pdflatex,sn-mathphys]{sn-jnl}

\usepackage{amsmath}
\usepackage{float}
\usepackage{hyperref}
\usepackage{dsfont}
\usepackage{xurl}
\usepackage{commath}

\jyear{2021}%

\theoremstyle{thmstyleone}%
\theoremstyle{thmstyletwo}%

\theoremstyle{thmstylethree}%

\begin{document}

\title[Frame by frame completion probability of an NFL pass]{Frame by frame completion probability of an NFL pass}


\author*[1]{\fnm{Gustavo Pompeu da} \sur{Silva}}\email{gustavopompeu@usp.br}

\author[1,2]{\fnm{Rafael de Andrade} \sur{Moral}}\email{Rafael.DeAndradeMoral@mu.ie}

\affil*[1]{\orgdiv{Departament of Exact Sciences}, \orgname{University of São Paulo}, \orgaddress{\street{Av. Pádua Dias, 11}, \city{Piracicaba}, \postcode{13418-900}, \state{São Paulo}, \country{Brazil}}}

\affil[2]{\orgdiv{Department of Mathematics and Statistics}, \orgname{Maynooth University}, \orgaddress{\city{Maynooth}, \postcode{10587}, \state{County Kildare}, \country{Ireland}}}


\abstract{American football is an increasingly popular sport, with a growing audience in many countries in the world. The most watched American football league in the world is the United States’ National Football League (NFL), where every offensive play can be either a run or a pass, and in this work we focus on passes. Many factors can affect the probability of pass completion, such as receiver separation from the nearest defender, distance from receiver to passer, offense formation, among many others. When predicting the completion probability of a pass, it is essential to know who the target of the pass is. By using distance measures between players and the ball, it is possible to calculate empirical probabilities and predict very accurately who the target will be. The big question is: how likely is it for a pass to be completed in an NFL match while the ball is in the air? We developed a machine learning algorithm to answer this based on several predictors. Using data from the 2018 NFL season, we obtained conditional and marginal predictions for pass completion probability based on a random forest model. This is based on a two-stage procedure: first, we calculate the probability of each offensive player being the pass target, then, conditional on the target, we predict completion probability based on the random forest model. Finally, the general completion probability can be calculated using the law of total probability. We present animations for selected plays and show the pass completion probability evolution.}




\maketitle

\section{Introduction}\label{sec1}

American football is a team sport played by two teams of eleven players on a rectangular field with goalposts at each end. The game is divided in plays; in each play one team is the offense and has possession of the ball, and the other is the defense. The offense tries to advance down the field by running or passing the ball, while the defense aims to stop the offense's advancement while trying to take control of the ball themselves. The most common ways of scoring points is by advancing the ball to the opponent's end zone for a touchdown, or by kicking the ball through the opponent's goalposts for a field goal. The team with most points at the end of the game wins.

The National Football League (NFL) is the most popular professional league of American football in the world. It is based in the United States and currently consists of 32 teams, divided in two conferences of four divisions each. The NFL is the most profitable professional sports league in the United States, having generated revenue of 15.26 billion U.S. dollars in 2019 \cite{statista2}. Each season is concluded with the Super Bowl, where the champions of each conference play against each other, and is one of the largest events of the year in the United States, with a growing audience around the world. Super Bowl LV, which was played on February 7th, 2021, had an average viewership of almost 100 million in the United States plus an estimated 30 to 50 million viewers around the world. Although it is still a small viewership when compared to the biggest events in the world like the FIFA World Cup Finals, it is a number that is growing every year \cite{statista}.

In this paper we focused on passing plays in the NFL, using data from the 2018 NFL season to model and predict probabilities of pass completion, in addition to estimating probabilities for each eligible player of the offense to be the target on every play.

We started by cleaning the data and creating the variables we were interested in including in our model (mostly distance measures) to obtain empirical probabilities of being the pass target for each of the offensive players in every frame. Using machine learning algorithms, we obtained the probabilities of pass completion given that a specific player was the target and then, through the law of total probability we estimated the probability of pass completion for the play as a whole \cite{zwillinger1999crc}.

In Section \ref{chap1}, we introduce the data set, along with the mathematical definition of the metrics we created. We also present an exploratory data analysis for context, and the statistical modeling tools used. In Section \ref{chap2}, we present and discuss our results. Finally, in Section \ref{chap3} we make our final considerations about our work, and draw conclusions and insights about how our results can be useful and how the methods can be improved further for future works.

\section{Materials and Methods}
\label{chap1}

The R software \cite{R} was used to read the data, build the models, generate the graphics and every other computational implementation needed. Many packages were used to obtain the results needed; to manipulate the data we used the \texttt{tidyverse} package \cite{tidy}, to easily implement the cross-validation for the models we used the \texttt{caret} package \cite{caret}, to fit the random forests we used the \texttt{randomForest} package \cite{rf},  and to create animations of the plays we used the \texttt{gganimate} package \cite{animate}. All R code utilized in this paper is made available at \url{https://github.com/gustavopompeu/NFLPassCompletion}.

\subsection{NFL data}

The data utilized in this paper was obtained from the Kaggle analytics competition NFL Big Data Bowl 2021 \cite{kaggle}, and is available at \url{https://www.kaggle.com/c/nfl-big-data-bowl-2021/data}. The competition used NFL's Next Gen Stats data that includes the position and speed of every player on the field during each play. The data contains tracking, play, game, and player information for all possible passing plays during the 2018 regular season, except from three games of week 1, for a total of $253$ games. Passing plays are considered to be the ones on which a pass was thrown, the quarterback was sacked, or any one of five different penalties was called (defensive pass interference, offensive pass interference, defensive holding, illegal contact, or roughing the passer). For each play, linemen (both offensive and defensive) data are not provided.

The data is hierarchical by nature, having game data, play data within each game and tracking data within each play. Besides that we also have player data. The utilized variables from each data level are shown below:

\begin{itemize}
    \item Game data: game identifier code, and the three-letter abbreviation codes of the home and visitor team;
    \item Player data: player identification number (unique across players), player name, and player position group (e.g. quarterback (QB), running back (RB), linebacker (LB), etc., totaling 8 categories);
    \item Play data: game identifier code, play identifier code, play description, game quarter (categorical, 1 to 5, with 5 representing overtime), down (categorical, 1 to 4), distance needed for a first down (in yards), three-letter abbreviation codes of possession team and which side of the field is the line-of-scrimmage, yard line at line-of-scrimmage, formation used by possession team (e.g. shotgun, wildcat, etc., totaling 7 categories), number of defenders in close proximity to line-of-scrimmage, number of pass rushers, dropback categorization of quarterback (e.g. designed rollout left, traditional, etc., totaling 7 categories), home and visiting team scores prior to the play (in points), time on clock of play (in minutes and seconds), NFL categorization of the penalties that occurred on the play, and outcome of the passing play (C: Complete pass, I: Incomplete pass, S: Quarterback sack, IN: Intercepted pass, totaling 4 categories);
    \item Tracking data: Player position along the long axis of the field (0 - 120 yards), player position along the short axis of the field, (0 - 53.3 yards), tagged play details (moment of ball snap, pass release, pass catch, tackle, etc., totaling 41 categories), player identification number, player name and jersey number, player position group (QB, RB, LB, etc.), team of corresponding player, frame identifier for each play (starting at 1), game identifier code, play identifier code, and direction to which the offense is moving (left or right).
\end{itemize}

The variable that indicates the direction to which the offense is moving was used to flip the coordinates $x$ and $y$ when the direction was left, so the plays always align with the direction of the offense's target end zone. A tutorial post on the Kaggle competition forum \cite{kaggletut} contained initial code to read and merge the databases and flip coordinates (\url{https://www.kaggle.com/tombliss/tutorial}).

\subsubsection{Data manipulation}

First and foremost, all the plays that presented some kind of problem in the database were removed from the data (for example, unusual pass plays like fake punts or fake field goals that don't have a specific offense formation would have a missing value for this variable, or plays that had clear problems such as not having tracking information for the ball or some players). All plays whose result was a sack were also removed, because although they are considered passing plays by the NFL, a pass does not actually take place. Plays that had a penalty in them were also removed because most of them would have missing values for several variables and we cannot be sure what the penalty is during plays, since the referees only announce the penalties after the play is over.

The variable of play description was very important. It describes what happened in the play, for example: ``(15:00) M.Ryan pass short right to J.Jones pushed ob at ATL 30 for 10 yards (M.Jenkins).''. This contains the name of the player that passed the ball, the player that received the pass and can have other characteristics of the play. From these descriptions, using string manipulation we were able to extract the name of the passer in every play and the name of the target in most of the plays. There were, however, some incomplete passes which did not state the name of the target on the play. In these cases the eligible receiver closest to the ball at the moment of the play being considered an incomplete pass was considered to be the target. Moreover, plays that were not meant to be attempted passes to a target (e.g. spikes -- when the quarterback simply throws the ball to the ground for the clock to stop, and throwaways -- when the quarterback is very pressured by the defense and throws the ball away to avoid a sack) could be easily identified because key words such as ``spiked'' or ``threw away'' would be present in the text. These plays were also removed from the data.

For the remaining plays we used the variable that tags the play details on every frame (moment of ball snap, pass release, pass catch, tackle, etc.) to determine in which one the forward pass begins and in which one there is an outcome to the play, such as completed pass, incomplete pass or interception. Finally, we filtered only the frames in between these events.

\subsection{Distance measures}
\label{distance}

It is noteworthy to mention that any field in the NFL is divided in yards, which is the standard distance measure in the NFL, so all the distances we calculated in this paper are also in yards.

The first distance measure we wanted to calculate was the distance of a point $(x_0,y_0)$ to the line created by the points $(x_1,y_1)$ and $(x_2,y_2)$, representing the movement of the ball or a player from one frame $t-1$ to another frame $t$. We refer to this distance as $d$.

Let $\boldsymbol{v}$ be a vector perpendicular to the line formed by $(x_1,y_1)$ and $(x_2,y_2)$, and given by
\begin{equation}
\label{eq1}
    \boldsymbol{v} = \begin{bmatrix} y_2-y_1\\ -(x_2-x_1) \end{bmatrix}.
\end{equation}
Then let $\boldsymbol{r}$ be a vector from the point $(x_0,y_0)$ to $(x_1,y_1)$:
\begin{equation}
\label{eq2}
    \boldsymbol{r} = \begin{bmatrix} x_1-x_0\\ y_1-y_0 \end{bmatrix}.
\end{equation}
We can calculate the distance $d$, which is given by projecting $\boldsymbol{r}$ onto $\boldsymbol{v}$ \cite{pointdist}
\begin{equation}
\label{eq3}
    d = \abs{\boldsymbol{\hat{v}} \cdot \boldsymbol{r}} = \frac{\abs{(x_2 - x_1)(y_1-y_0) - (x_1-x_0)(y_2-y_1)}}{\sqrt{(x_2-x_1)^2+(y_2-y_1)^2}}.
\end{equation}

\begin{figure}[!htbp]
\centering
\includegraphics[width=0.4\textwidth]{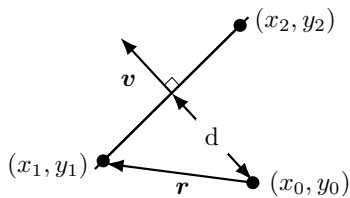}
\caption{Visual representation of the Euclidean point to line distance $d$, given by Equation \ref{eq3}}
\label{fig2}
\end{figure}

In Figure \ref{fig2} we visualize the points and the vectors used to calculate the distance $d$. In R we created a function to calculate and return the distance $d$ when informing the coordinates of the three points, using the formula given by Equation \ref{eq3}.

To calculate the distance between two points on a plane, we used the Euclidean distance between two points, which for any set of points $(x_0,y_0)$ and $(x_1,y_1)$ is given by

\begin{equation}
\label{eq4}
    h = \sqrt{(x_1-x_0)^2 + (y_1-y_0)^2}.
\end{equation}

In R, $h$ was calculated through the function \texttt{PointDistance} from package \texttt{raster} \cite{raster}, which uses the formula in Equation \ref{eq4}. Finally, we also define the metric $b$ as the difference of distances for a player or for the ball between frames $t$ and $t-1$:
\begin{equation}
\label{eq5}
    b = h_t - h_{t-1}.
\end{equation}

\subsection{Statistical modeling}

We mainly separated our work into two big modeling issues; one to determine probabilities of the offensive players to be the target of the play, and the other to determine the probabilities of the pass to be completed given that a specific player is the target.

\subsubsection{Target prediction}
\label{231}

To calculate distances between players and the ball, we used the variables that describe their coordinates on the field in each frame, and applied them to the equations described in Section \ref{distance}. There were some cases where the ball coordinates in the previous frame were identical to the current frame; in these situations we considered the coordinates of the previous frame that had different coordinates to calculate the distance from point (player) to line (ball direction) described in Equation \ref{eq3}. We refer to this distance as $d^{(1)}$. In the case of the location of a player being exactly on top of the line created by the ball's coordinates we set the distance to $0.01$ to avoid issues of having a distance of $0$ in the algorithm we describe below.

In some plays the ball was still going backwards on the first frames despite the frames being after the event of a ``pass forward", most likely because of an error in the database or because the passer was moving backwards when releasing the ball. In these cases we discarded the first few frames until the ball really started moving forward.

From Equation \ref{eq5}, we call $d^{(3)}$ the distance $b$, which represents the distance difference between players and ball from previous to current frame, resulting in a negative number if the distance from a player to the ball decreased in current frame, and positive number if the distance increased. A characteristic of this distance is that it can have negative values, so we created a standardised version $d^{(2)}$ such that $d^{(2)} \geq 1$, defined as

\begin{equation}
\label{eq6}
    d^{(2)}_{ij} = \begin{cases}
      d^{(3)}_{ij} + 1 + \abs{\min(\boldsymbol{d}^{(3)}_j)} & \text{if } \min(\boldsymbol{d}^{(3)}_j) > 0 \\
      d^{(3)}_{ij} + 1 - \min(\boldsymbol{d}^{(3)}_j) & \text{otherwise}
    \end{cases},
\end{equation}
where $d^{(2)}_{ij}$ and $d^{(3)}_{ij}$ represents the distance $d^{(2)}$ and $d^{(3)}$ for player $i$ on frame $j$ respectively, and $\boldsymbol{d}^{(3)}_j$ is the vector of all observations of $d^{(3)}$ for frame $j$. The use of $d^{(2)}$ avoids problems arising from values equal to or close to zero.

We calculated the probability of a player $i$ being the target for every frame $j$ considering each frame independent, even those in the same play. The rationale for this is that what was used to determine these probabilities were only the aforementioned distances, and since from one frame to another on the same play these distances don't change much, the probabilities for the same player on the same play end up following a natural dependence pattern. Using the distance variables $d^{(1)}$ and $d^{(2)}$, we calculated an empirical probability of player $i$ being the pass target $T$, given the information from frame $j$, from
\begin{equation}
\label{prob1}
    P_k(T = i \mid j) = \frac{\min(\boldsymbol{d}^{(k)}_j)}{d^{(k)}_{ij}} \times \frac{1}{\sum_i^n \frac{\min(\boldsymbol{d^{(k)}_j})}{d^{(k)}_{ij}}} = \frac{1}{d^{(k)}_{ij} \times \sum_{i=1}^n (d^{(k)}_{ij})^{-1}},
\end{equation}
where $k=1,2$, $i=1,\ldots,n$ represents the players who are the potential pass targets of the play in frame $j$, $d_{ij}^{(k)}$ is the distance measure $d^{(k)}$ of player $i$ in frame $j$, and $\boldsymbol{d}_j^{(k)}$ represents the vector of values of $d^{(k)}$ for all potential targets on frame $j$. This formula guarantees that the sum of the probabilities will be $1$ for every frame, and for example, considering $k=1$, guarantees that the closest player to the line projected by the ball will have the highest probability of being the target, while when $k=2$, the player that got closer to the ball from previous to current frame will have the highest probability of being the target.

To increase the accuracy of our method, we combined probabilities based on $d^{(1)}$ and $d^{(2)}$, and wrote this combination as a function of a weight $W$:
\begin{equation}
\label{weight}
    f(W) = W P_1(T = i \mid j) + (1-W) P_2(T = i \mid j),
\end{equation}
where $0\leq W\leq 1$. The objective was to give more importance to one measure or the other in each frame, depending on different  characteristics of the plays. If we consider that these two metrics have the same importance in every situation we will have $W=0.5$.

We used four different approaches to determine these weights, considering the order statistics of a metric for one player per frame in the whole database. Let $d^{(4)}$ be the Euclidean distance $h$ between the players and the ball. The four weights used were:

\begin{itemize}
    \item $W^{(1)}$, based on the $d^{(3)}$ of the player with the lowest $d^{(1)}$;
    \item $W^{(2)}$, based on the $d^{(2)}$ of the player with the lowest $d^{(1)}$;
    \item $W^{(3)}$, based on the $d^{(4)}$ of the player with the lowest $d^{(1)}$;
    \item $W^{(4)}$, based on the $d^{(4)}$ of the player with the lowest $d^{(2)}$.
\end{itemize}

The values of $W^{(r)}, r=1,2,3,4,$ are the same for every player $i$ on the same frame, but different for every frame $j$. Therefore we use the indexing $W^{(r)}_j$ to refer to weight of type $r$ calculated for frame $j$. To represent the values of these weights mathematically, we define the matrix

\begin{equation}
\label{matrixD}
    \boldsymbol{\mathds{D}}^{(k)}  = 
\begin{bmatrix}
    d_{11}^{(k)} & d_{12}^{(k)} & d_{13}^{(k)} & \dots  & d_{1m}^{(k)} \\
    d_{21}^{(k)} & d_{22}^{(k)} & d_{23}^{(k)} & \dots  & d_{2m}^{(k)} \\
    \vdots & \vdots & \vdots & \ddots & \vdots \\
    d_{n1}^{(k)} & d_{n2}^{(k)} & d_{n3}^{(k)} & \dots  & d_{nm}^{(k)}
\end{bmatrix},
\end{equation}
with $d_{ij}^{(k)}$ representing the distance measure $d^{(k)}$ of player $i=1,...,n$ in frame $j=1,...,m$, where $m=203$,148 represents the total number of frames in all of our database. Let $\boldsymbol{\mathds{Z}}^{(s)}$ ($s= 1,...,4$ same as index $k$) be a matrix whose elements are
\begin{equation}
\label{matrixK}
    z_{ij}^{(s)} = \begin{cases}
      1 & \text{if } \min_i(\boldsymbol{d}_j^{(s)}) = d_{ij}^{(s)}\\
      0 & \text{otherwise}
    \end{cases},
\end{equation}
where $\boldsymbol{d}_j^{(s)}$ is the $j$-th column of $\boldsymbol{\mathds{D}}^{(s)}$. We also define
\begin{equation}
\label{dk}
    \boldsymbol{d}_*^{(k)} = \text{diag}\{(\boldsymbol{\mathds{D}}^{(k)})^\top  \boldsymbol{\mathds{Z}}^{(s)}\},
\end{equation}
where $\text{diag}(\mathbf{X})$ represents the main diagonal of the square matrix $\mathbf{X}$.

Now let $\boldsymbol{U}$ be the ordered vector of dimension $u$ containing the unique values in $\boldsymbol{d}_*^{(k)}$, and $\boldsymbol{a}$ be a vector of the same dimension $u$ such that the $l$-th element of $\boldsymbol{a}$ is $a_l = \frac{l-1}{u-1}$, then
\begin{equation}
\label{wj}
    W^{(r)}_j = a_l, \ \text{for the} \ l \ \text{corresponding to the match} \  U_l = d_{*j}^{(k)}.
\end{equation}

\noindent To make Equation \ref{wj} more clear, suppose we have 7 frames and $\boldsymbol{d}_*^{(k)} = \{3,3,2,6,6,1,8\}$. Then $\boldsymbol{U} = \{1,2,3,6,8 \}$ and $\boldsymbol{a} = \{\frac{0}{4}, \frac{1}{4}, \frac{2}{4}, \frac{3}{4}, \frac{4}{4} \}$, with $\boldsymbol{W}^{(r)} = \{\frac{2}{4}, \frac{2}{4}, \frac{1}{4}, \frac{3}{4}, \frac{3}{4}, \frac{0}{4}, \frac{4}{4} \}$, because we are seeking the $l$-th element of $\boldsymbol{a}$ for the $l$ that corresponds to the match $U_l = d_{*j}^{(k)}$.

For each different weight, the values of $k$ and $s$ in these equations will depend on which distance measures the weights are based on. For $W^{(1)}$ we have $k=3$ and $s=1$ because it is based on the $d^{(3)}$ of the player with the lowest $d^{(1)}$. Then for $W^{(2)}$ we have $k=2$ and $s=1$, for $W^{(3)}$ we have $k=4$ and $s=1$, and finally for $W^{(4)}$ we have $k=4$ and $s=2$.

Substituting $W^{(r)}_j$ in Equation \ref{weight} we get the probabilities of every player being the target on frame $j$ based on weights of type $r$. In practice, what this means for weight $W^{(1)}$ is that the lower the $d^{(3)}$ of the player with the lowest $d^{(1)}$ is in a frame, we give more importance to the probability $P_1(T=i \mid j)$, from Equation \ref{prob1}, which means that if the player closest to the line projection of the ball is also getting much closer to the ball itself, we will give much more importance to the probability calculated from the distance of player to line projection of the ball. Similarly, the larger this distance is, we give more importance to the probability $P_2(T=i \mid j)$, which means that if the player closest to the line projection of the ball is actually getting farther away from the ball itself, we will give much more importance to the probability calculated from the distance difference between the current and previous frames. The same line of thought applies for the other weights, but for $W^{(4)}$ it is inverted, meaning that in Equation \ref{weight}, for this type of weight, we have to use $f(1-W^{(4)})$. This is because it is the only weight for which we calculate a distance for the player with the lowest $d^{(2)}$ and not the lowest $d^{(1)}$.

Now that we have a probability of every player being the target in every frame of every play, if we consider the player with highest probability in each frame to be the predicted target, we can calculate the accuracy of our method. We do so using each of the proposed weights. A problem with these probabilities is that even in situations where a player clearly does not have a chance to be the target anymore, the probability associated to this player will not be zero. To fix some of the most obvious cases of this problem, we made an adjustment. We considered that a player is very close to ball when he has $d^{(1)}$ and $d^{(4)}$ less than $2$ yards. When this happens, the probabilities of all the other players that are not within the same distances are added to his probability, and theirs are set to zero. In the extreme rare cases that there are more than one player very close to the ball, the probabilities are ``transferred'' to the one that has the highest probability of being the target. Table \ref{tableacc} shows the accuracy before and after applying the aforementioned adjustment.

\begin{table}[!htbp]
\centering
\caption{Accuracy of target prediction using different weights, before and after the adjustment of transferring probabilities when there is at least one player very close to the ball. EW represents equal weights ($W=0.5$ in Equation \ref{weight})}
\label{tableacc}
\begin{tabular}{ccc}
\hline
\textbf{Weight} & \textbf{Accuracy before adjustment} & \textbf{Accuracy after adjustment} \\ \hline
EW        & $81.25\%$  & $82.67\%$          \\ 
$W^{(1)}$              & $84.15\%$  & $85.48\%$          \\ 
$W^{(2)}$              & $85.66\%$  & $86.68\%$           \\ 
$W^{(3)}$              & $84.34\%$  & $85.23\%$           \\ 
$W^{(4)}$              & $81.06\%$  & $81.89\%$          \\ \hline
\end{tabular}
\end{table}

Of all plays in the data, the number of frames analysed varied from $1$ to $46$, but was mostly concentrated below $20$. $75\%$ of the plays consisted of $16$ or fewer frames, and $95\%$ of the plays had $25$ or fewer. With this information, we decided to analyse the accuracy considering the frames, both from the beginning of plays and from the end of plays, to see specifically whether the tested weights would have different performances in these situations.

\begin{figure}[!htbp]
  \centering
  \includegraphics[width=0.9\textwidth]{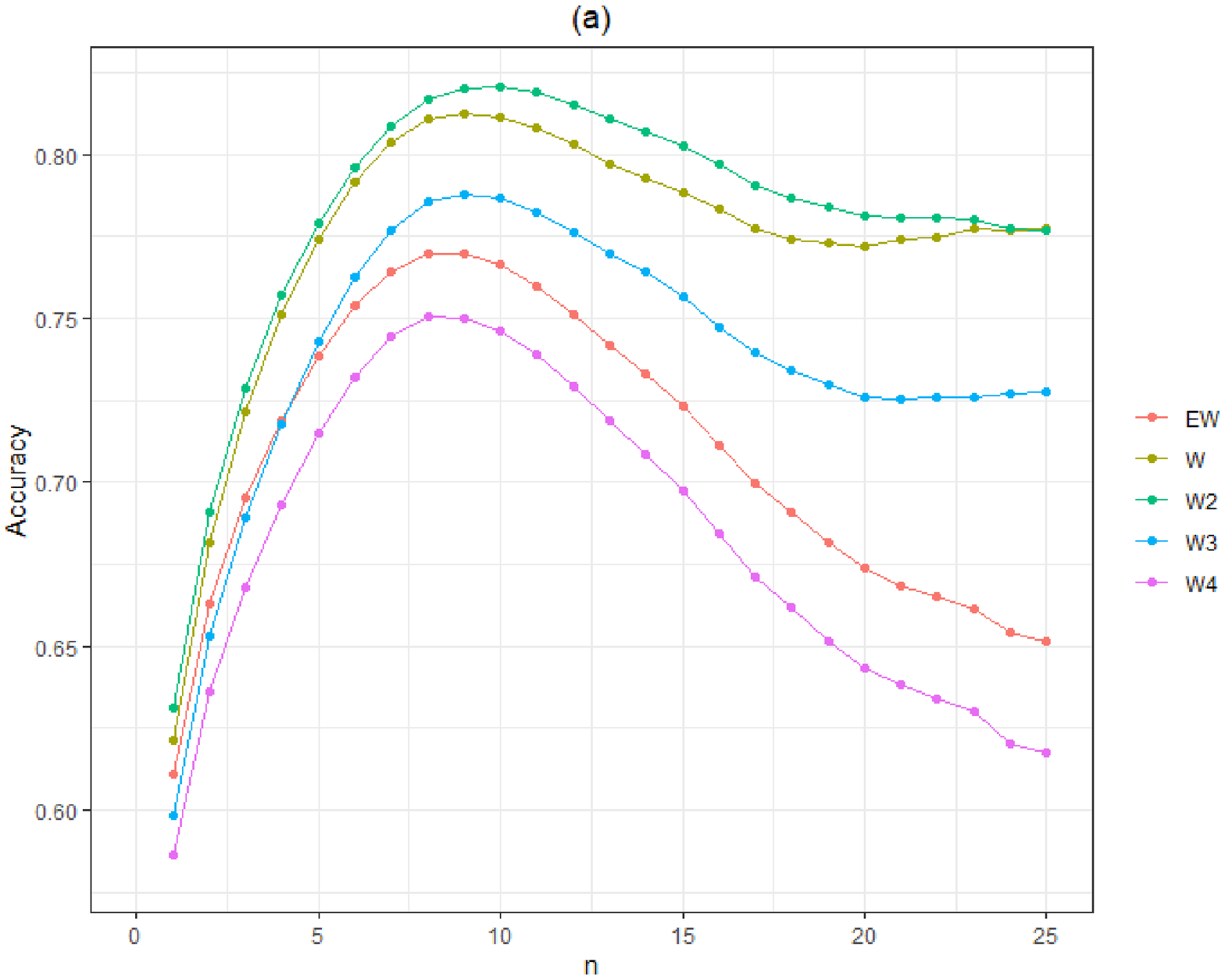}
  \includegraphics[width=0.9\textwidth]{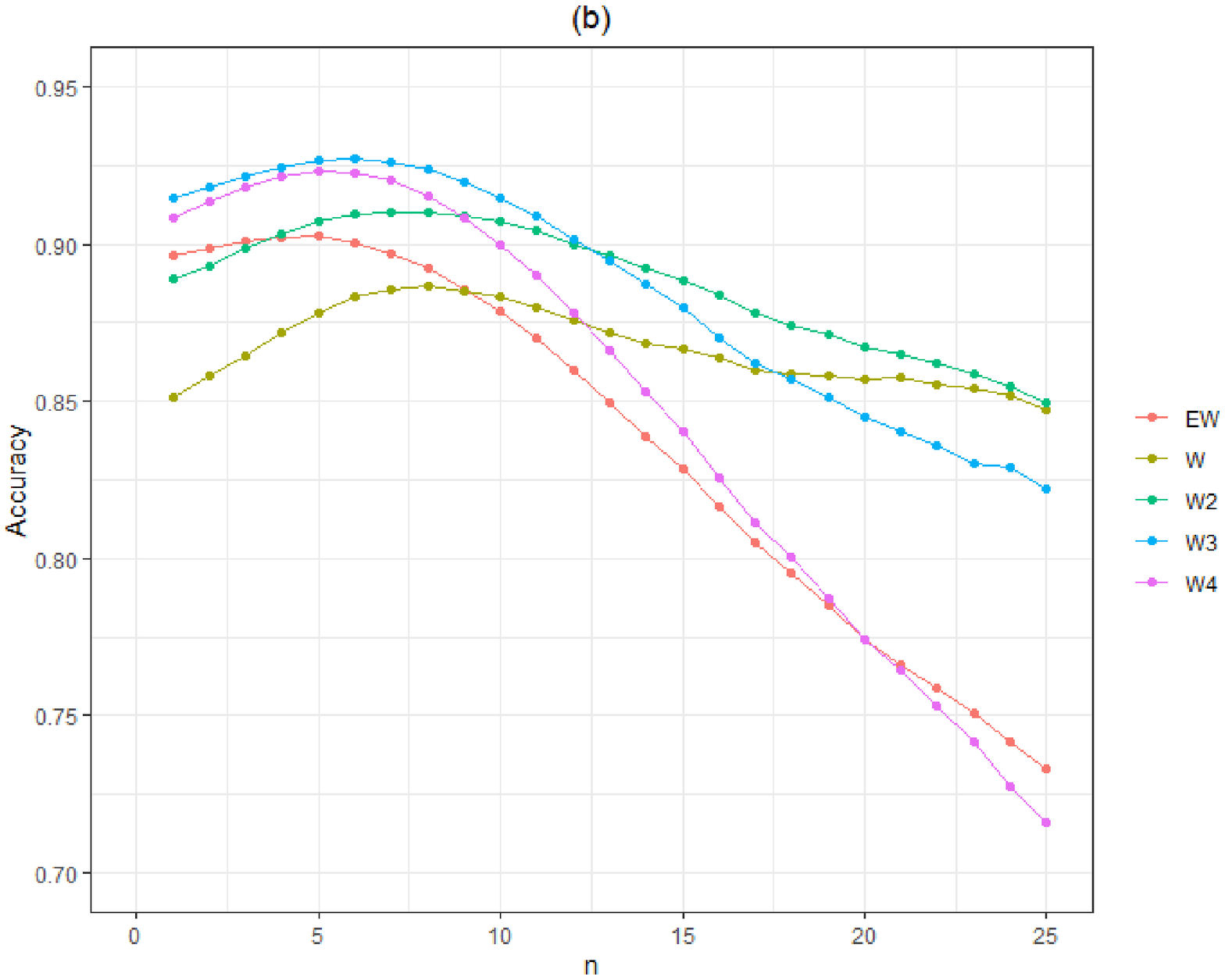}
  \caption{Accuracy of Target prediction in (a) first or (b) last $n$ frames of plays, i.e., proportion of frames that the player with highest probability of being the target really was the target}
  \label{target}
\end{figure}

In Figure \ref{target}, we can see that looking from the beginning of the plays, weight $W^{(2)}$ seems to provide the best accuracy for all frames. But when we look at the last frames of the plays, we see that weight $W^{(3)}$ has a better accuracy in the last $12$ frames.

This characteristic propelled us to create a new weight combining $W^{(2)}$ and $W^{(3)}$, giving more importance to $W^{(2)}$ in the beginning of plays but changing to give more importance to $W^{(3)}$ as the play extends. We used a logistic model to create these new weights. The values to evaluate the model were $1$ to $46$, which are the minimum and maximum number of frames found in a play, with asymptote $1$, inflection point of the curve being the mean of number of frames in the plays, which was $13.34183$, and scale parameter estimated to be $2.57$ via grid search with accuracy as our target function. We called these new weights $W^{(2,3)}$, mathematically defined as
\begin{equation}
\label{w23}
    W^{(2,3)}_t = \frac{1}{1+e^\frac{(13.34183 - t)}{2.57}},
\end{equation}
with $t=1,...,46$ representing the number of the frame in a play. The difference between $t$ and $j$ is that $t$ is specific for each play, variating from $1$ to $46$ (maximum of frames evaluated in a play), and $j$ is a generic count of frames on the whole database, from $1$ to 203,148 (total number of frames on the database).

Therefore, the final formula we used to calculate the probability of a player $i$ to be the target of a play in frame $j$ is
\begin{equation}
\label{probw}
    P(T = i \mid j, t, \boldsymbol{\mathds{D}}^{(1)}, \boldsymbol{\mathds{D}}^{(2)}, \boldsymbol{\mathds{D}}^{(4)}) = W^{(2,3)}_t f(W^{(3)}_j) + (1-W^{(2,3)}_t) f(W^{(2)}_j).
\end{equation}
This new and final way to compute the probabilities achieved an accuracy of $86.92\%$, which was better than the ones obtained through either $W^{(2)}$ or $W^{(3)}$, as expected.

\subsubsection{Completion probability}
\label{232}

Now that we have probabilities of each eligible offensive player being the target, we can calculate the conditional probability of the pass to be completed given that a player is the target. Using the law of total probability we can then compute the probability of a pass being completed for each frame of every play. We write
\begin{equation}
    \label{totalp}
    P(C) = \sum_{i=1}^n P(C \mid T = i)P(T = i),
\end{equation}
where $P(C)$ is the completion probability, $n$ is the number of players that can be the target on a given play, $P(T=i)$ is the probability of player $i$ being the target, calculated through Equation \ref{probw}, and $P(C \mid T=i)$ is the completion probability given that player $i$ is the target. The computation of $P(C \mid T=i)$ is the focus of this section.

The response variable for the models we tested was a binary variable with two levels: complete or incomplete pass. Pass interceptions were considered incomplete passes. We used $32$ explanatory variables, listed below. As mentioned earlier, all distance measures are given in yards:

\begin{itemize}
    \item Play data:
    \begin{itemize}
        \item Game quarter (factor, with levels 1 to 5, the last one representing overtime);
        \item Down (factor, with levels 1 to 4);
        \item Distance needed for a first down (numeric);
	    \item Formation used by possession team (factor, with 7 categories);
        \item Number of defenders in close proximity to line-of-scrimmage (numeric);
        \item Number of pass rushers (numeric);
	    \item Dropback categorization of quarterback (factor, with 7 categories);
	    \item Time on clock of play, in seconds (numeric);
	    \item Yard line at line-of-scrimmage from 1-99 (numeric);
	    \item Offensive team score prior to the play (numeric);
        \item Defensive team score prior to the play (numeric);
        \item Indicator if the offensive team is playing at home (logical);
        \item Distance from passer to given target at the moment of pass (numeric);
    \end{itemize}
    \item Player data:
    \begin{itemize}
        \item Player position of the given target (WR, RB, TE, etc.) (factor, with 5 categories);
        \item Player position of the closest defensive player (LB, DB, S, etc.) (factor, with 5 categories);
        \item Player position of the second closest defensive player (LB, DB, S, etc.) (factor, with 5 categories).
    \end{itemize}
    \item Frame data:
    \begin{itemize}
        \item Distance from given target to line projection created by the ball (numeric);
        \item Distance from given target to the ball (numeric);
        \item Distance difference in current and previous frame from given target to the ball (numeric);
        \item Distance from closest defensive player to line projection created by the ball (numeric);
        \item Distance from closest defensive player to the ball (numeric);
        \item Distance difference in current and previous frame from closest defensive player to the ball (numeric);
        \item Distance from second closest defensive player to line projection created by the ball (numeric);
        \item Distance from second closest defensive player to the ball (numeric);
        \item Distance difference in current and previous frame from second closest defensive player to the ball (numeric);
        \item Distance from closest defensive player to line projection created by the given target (numeric);
        \item Distance from closest defensive player to the given target (numeric);
        \item Distance difference in current and previous frame from closest defensive player to the given target (numeric);
        \item Distance from second closest defensive player to line projection created by the given target (numeric);
        \item Distance from second closest defensive player to the given target (numeric);
        \item Distance difference in current and previous frame from second closest defensive player to the given target (numeric);
        \item Distance from given target to the nearest sideline (numeric).
    \end{itemize}
\end{itemize}

The closest defensive player mentioned was considered to be the closest defensive player to the line projection of the given target, and the second closest defensive player is the closest defensive player to the given target in Euclidean distance. When the same player is the closest by both metrics, we considered the second closest defensive player to be the second closest in Euclidean distance.

\begin{figure*}[!htbp]
  \centering
  \includegraphics[width=0.45\textwidth]{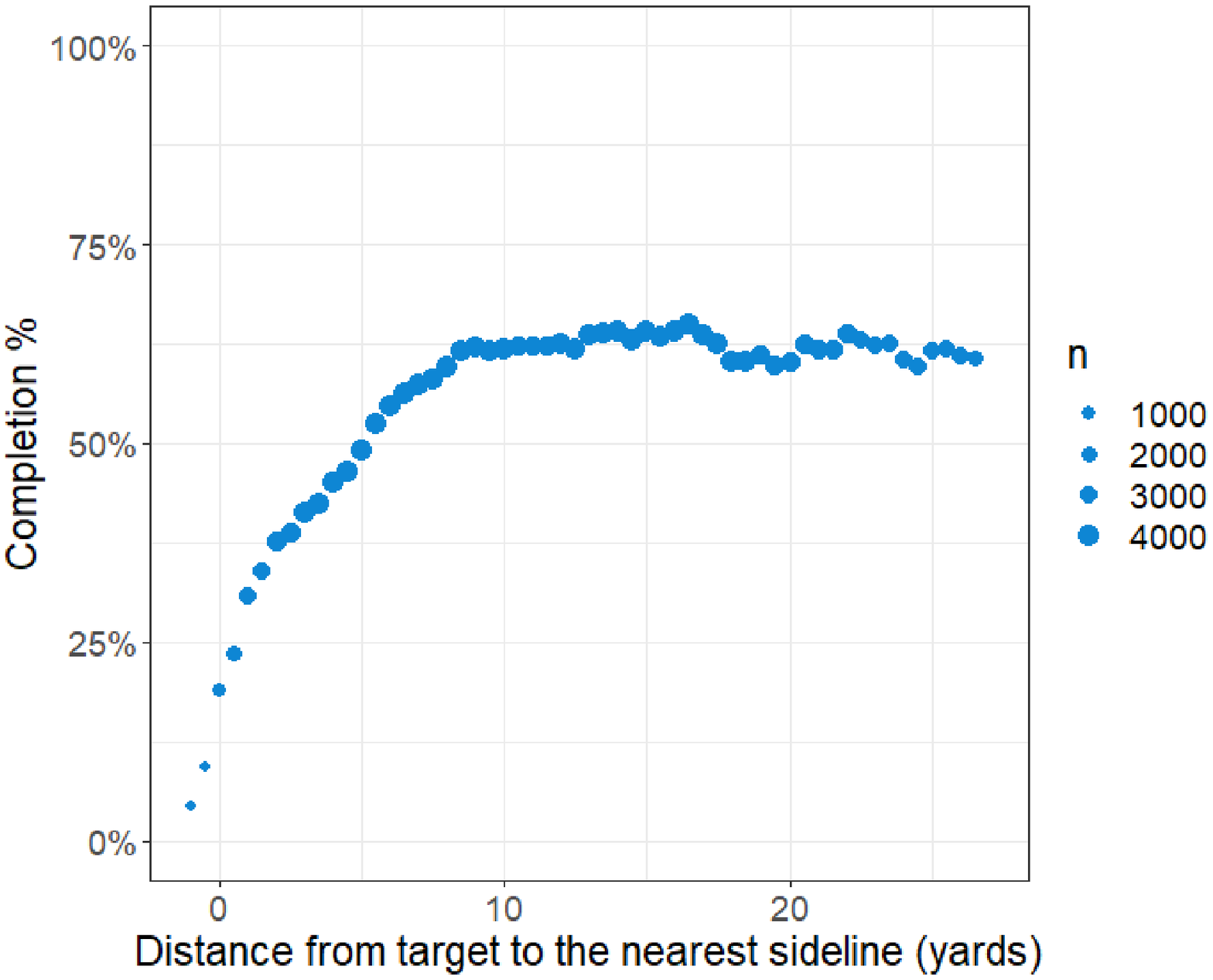}
  \includegraphics[width=0.45\textwidth]{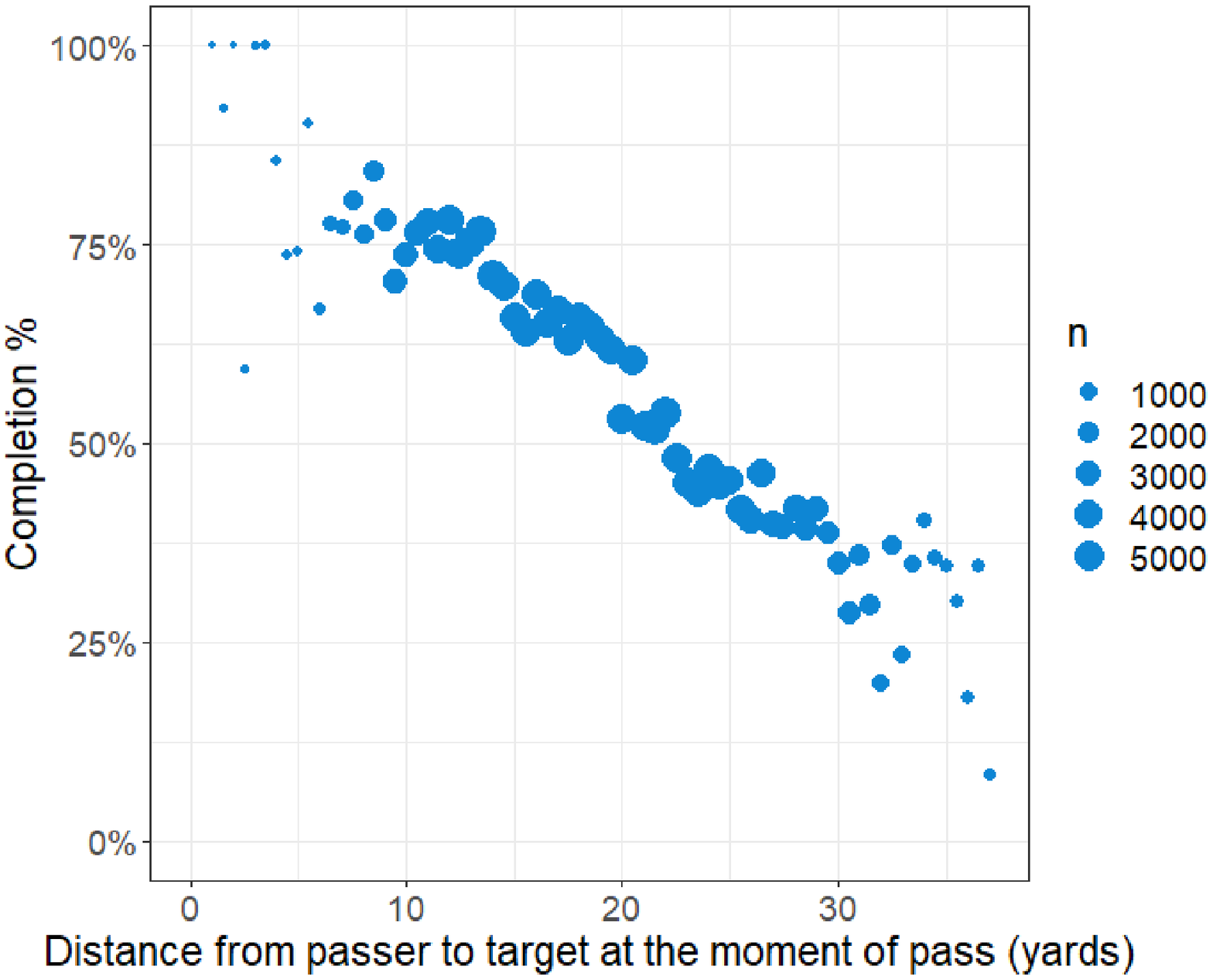}
  \includegraphics[width=0.45\textwidth]{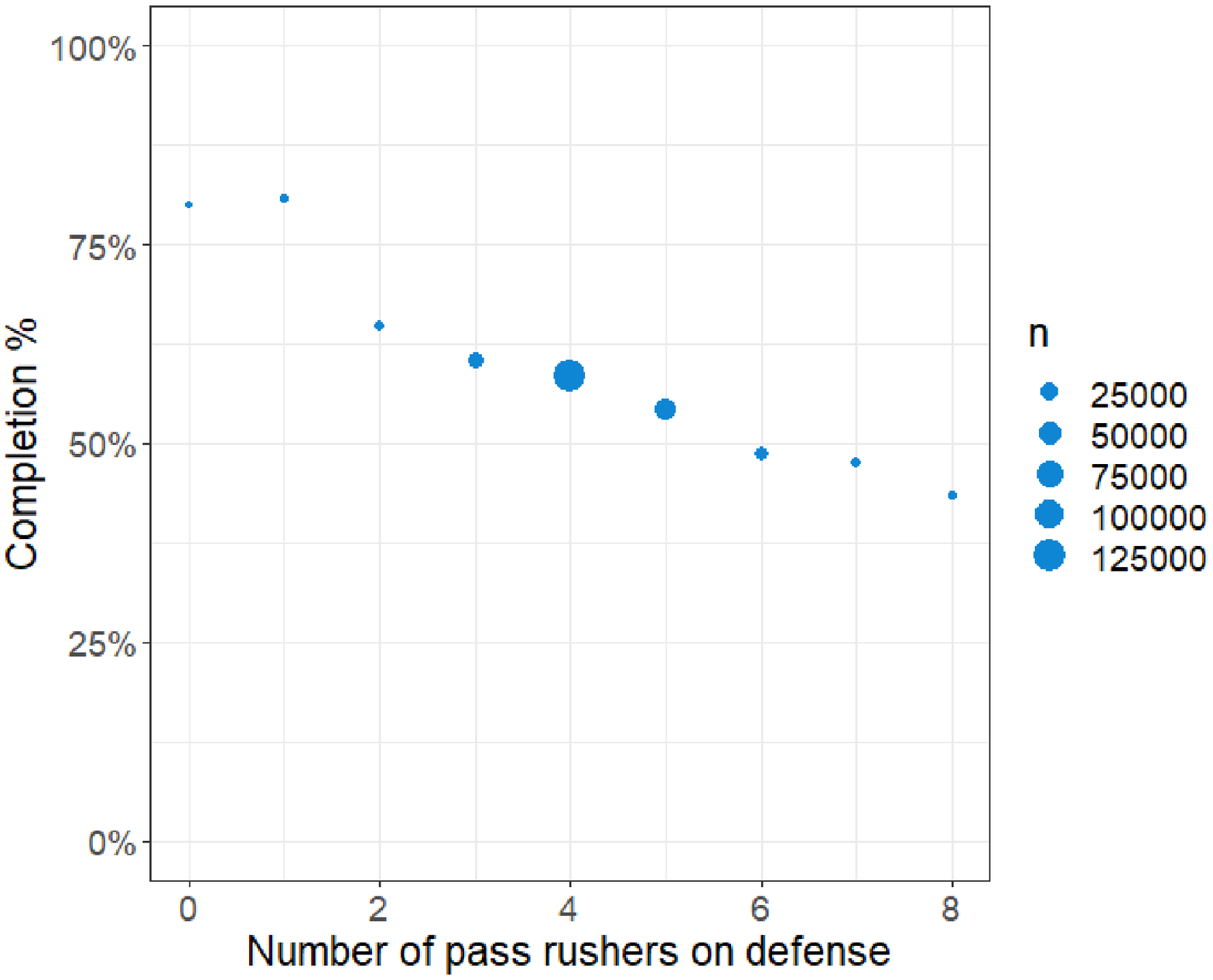}
  \includegraphics[width=0.45\textwidth]{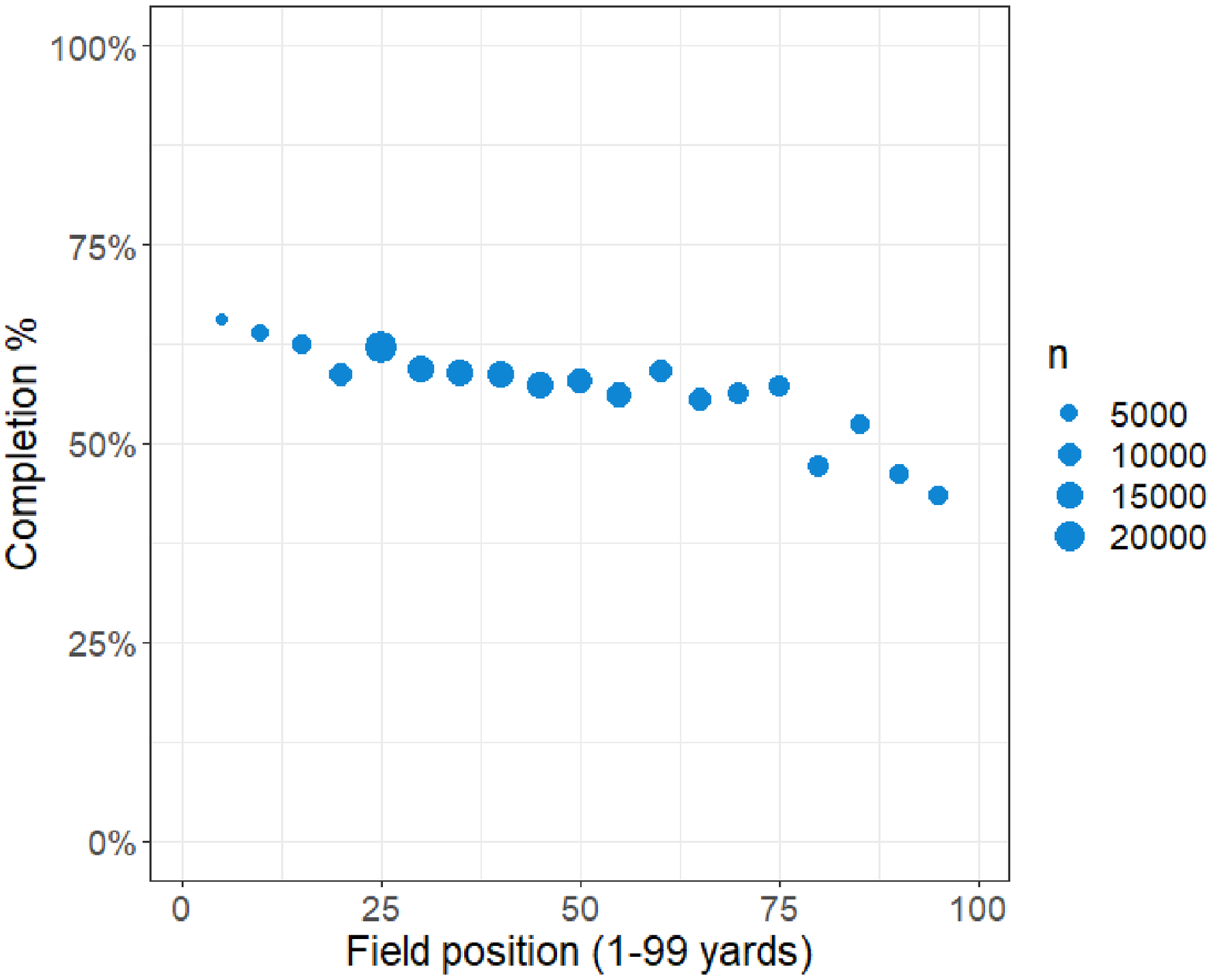}
  \includegraphics[width=0.45\textwidth]{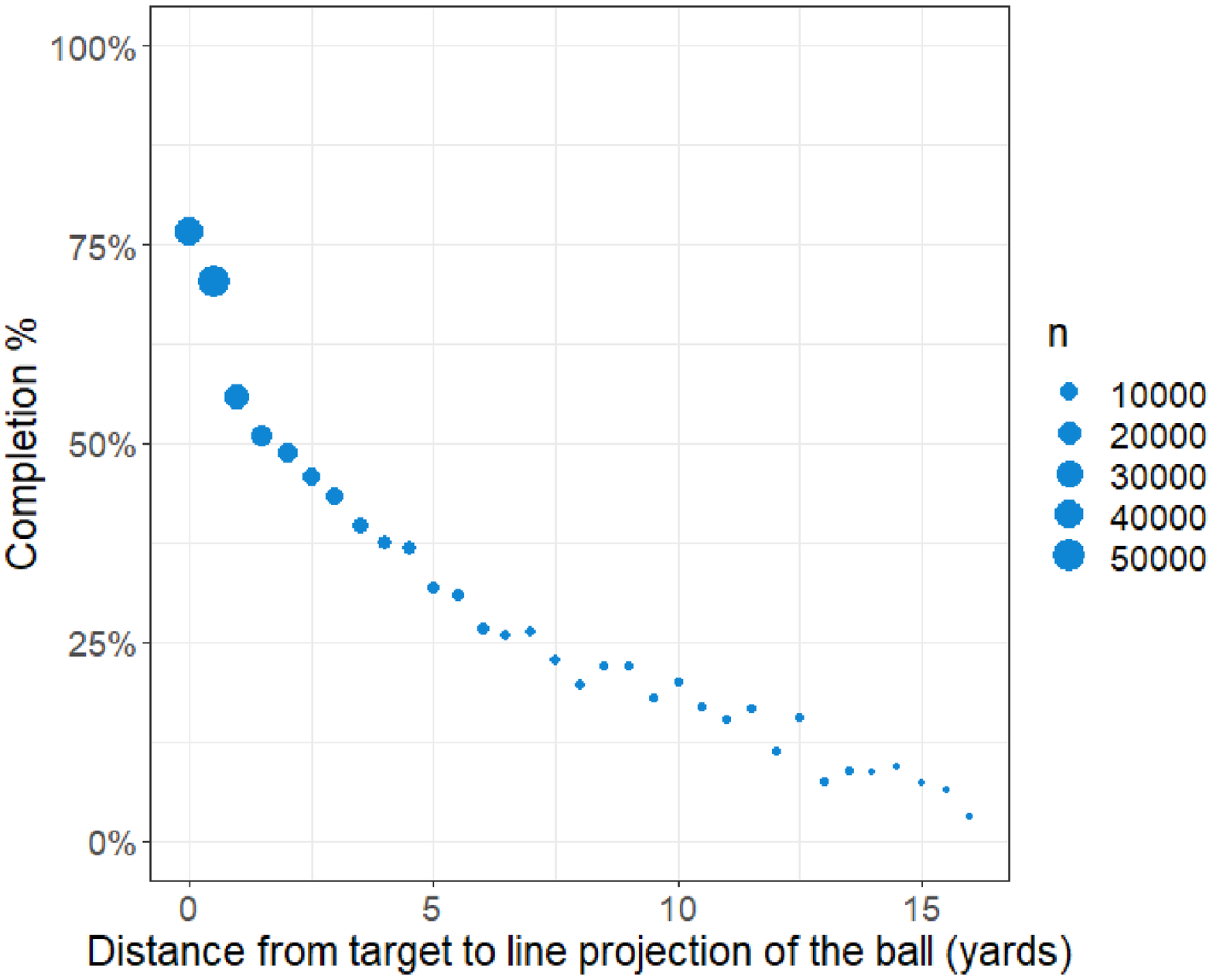}
  \includegraphics[width=0.45\textwidth]{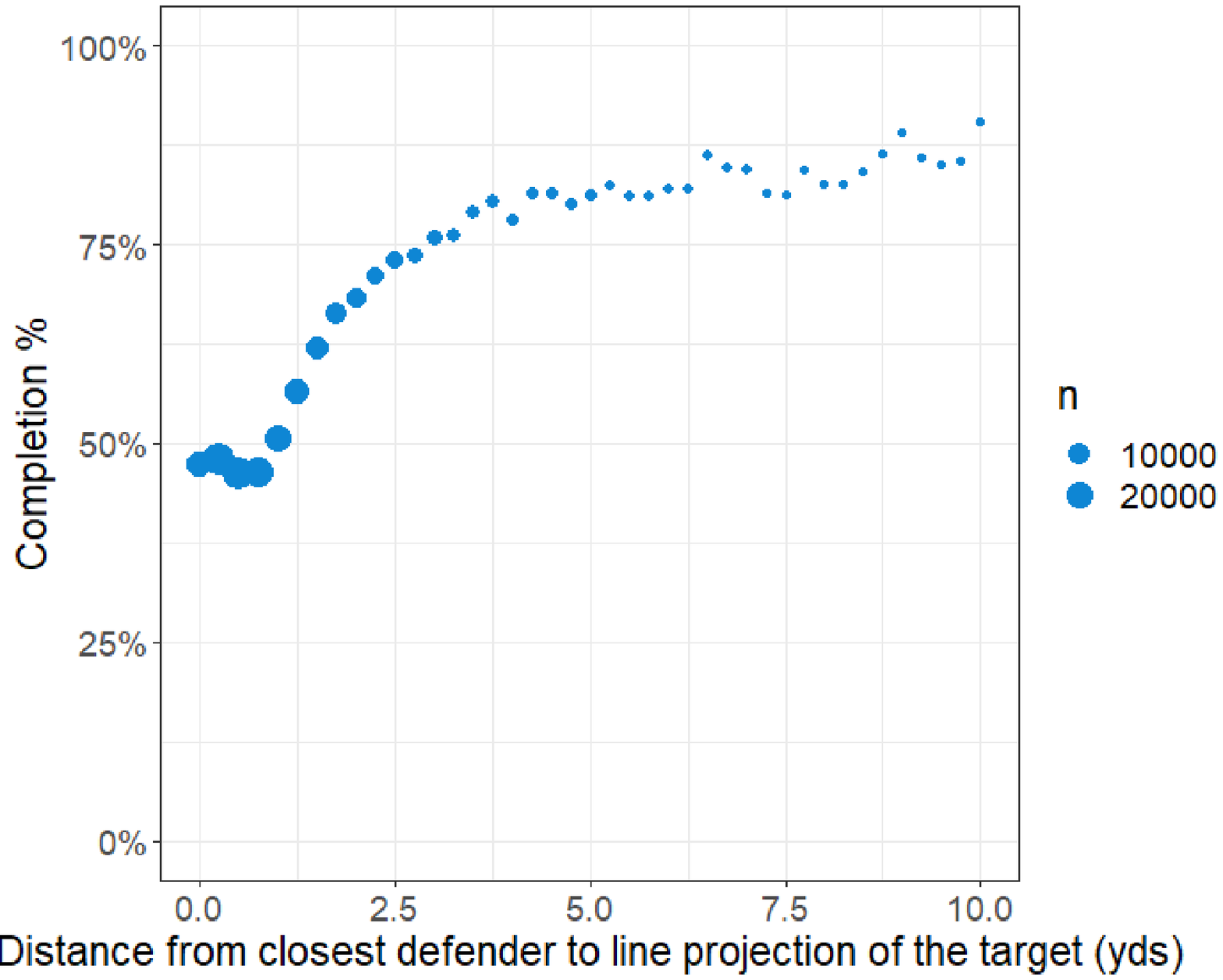}
  \caption{Plots of six different variables versus the completion percentage of passes, showing the correlation between explanatory and response variables. The continuous variables displayed on the $x$ axis were discretized in intervals, and the size of the points correspond to the number of frames belonging to each interval}
  \label{plots}
\end{figure*}

In Figure \ref{plots}, we demonstrate the correlation between some of these explanatory variables and the completion or incompletion of a pass, where the $y$ axis in all plots represent the completion percentage, i.e., the percentage of frames corresponding to the metric in the $x$ axis where the result of the play was a completed pass.

We tested the following different models: Random Forest, Binomial Regression with Logit, Probit and complementary-log-log links, and Linear and Quadratic Discriminant Analysis. The objetive was to know which one would have the best overall performance. To test these models we used Leave Group Out Cross-Validation (LGOCV) with $5$ or $10$ folds, using plays as the grouping factor, meaning that all frames within a play were held out in one of the folds. To test the models, only data from the real target of the plays were used, meaning we have one observation per frame.

The performance metric chosen to evaluate the models was the area under the Receiver Operating Characteristic (ROC) curve \cite{roc}, computed with the trapezoidal rule \cite{proc}, abbreviated in Table \ref{table1} as AUC, as it is one of the most widely used metrics for evaluation of binary classification problems. The AUC is a robust overall measure to evaluate the performance of score classifiers because its calculation relies on the complete ROC curve and thus involves all possible classification thresholds \cite{hanley}.

\begin{table}[!htbp]
\centering
\caption{Comparison between all the different models tested, sorted by AUC (area under the ROC curve) in descending order, as well as computational time taken to run the cross-validation procedures. BR = Binomial Regression, LDA = Linear Discriminant Analytis, QDA = Quadratic Discriminant Analysis,``cloglog" is the complementary log-log link}
\label{table1}
\begin{tabular}{cccc}
\hline
\textbf{Method}    & \textbf{Folds} & \textbf{AUC} & \textbf{Time}        \\ \hline
Random Forest      & 10             & 0.8829             & $\sim$3.88 hours \\
Random Forest      & 5              & 0.8825             & $\sim$1.78 hours \\
BR (logit link)   & 10             & 0.7874             & 2.00 mins        \\
BR (logit link)   & 5              & 0.7877             & 56.16 secs        \\
BR (probit link)  & 10             & 0.7861             & 5.04 mins        \\
BR (probit link)  & 5              & 0.7864             & 2.08 mins        \\
LDA                & 10             & 0.7840             & 1.09 mins        \\
LDA                & 5              & 0.7843             & 32.35 secs        \\
BR (cloglog link) & 10             & 0.7814             & 6.30 mins        \\
BR (cloglog link) & 5              & 0.7816             & 2.94 mins        \\
QDA                & 10             & 0.7487             & 38.58 secs        \\
QDA                & 5              & 0.7462             & 21.02 secs        \\ \hline
\end{tabular}
\end{table}

\begin{figure}[!htbp]
  \centering
  \includegraphics[width=0.9\textwidth]{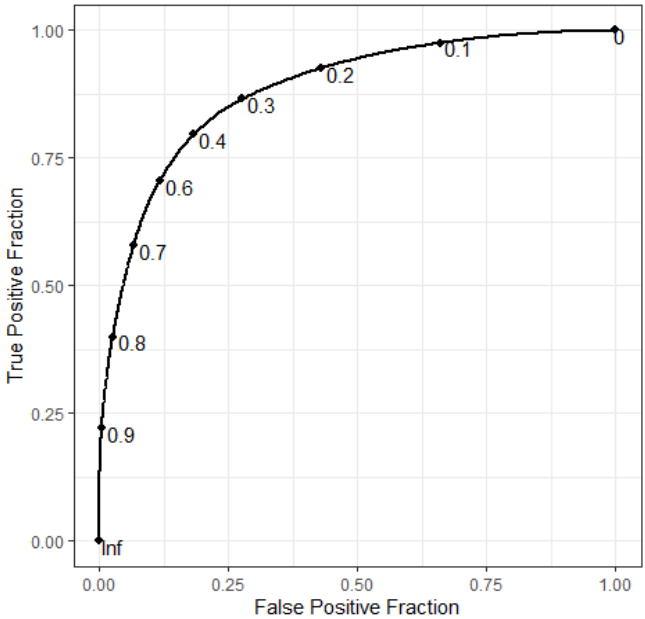}
  \caption{ROC curve for the Random Forest model, based on 10-fold leave-group-out cross-validation}
  \label{roc}
\end{figure}

For each tree within the Random Forest algorithms we allowed subsets containing between 5 to 20 variables to be chosen at every split, and we obtained the best results based on the AUC when this number was 15. This corresponds to the \texttt{mtry} argument within the function \texttt{randomForest()} from package \texttt{randomForest} on R. The values shown in Table \ref{table1} for the method Random Forest refer to the value of $\texttt{mtry}=15$. We can see that the performance of this metric does not change much for the same method when comparing between 5 or 10-fold cross-validation, as expected. However, there is a large difference between some of the methods, and the Random Forest models were vastly superior to the other methods, but were associated to larger computational burden (although not that large, since four hours is not a long time considering such a big dataset). We therefore chose the Random Forest model to draw our results and analysis from. The ROC curve of the Random Forest model based on the 10-fold leave-group-out cross-validation can be seen in Figure \ref{roc}.

\begin{figure}[!htbp]
\centering
\includegraphics[width=0.9\textwidth]{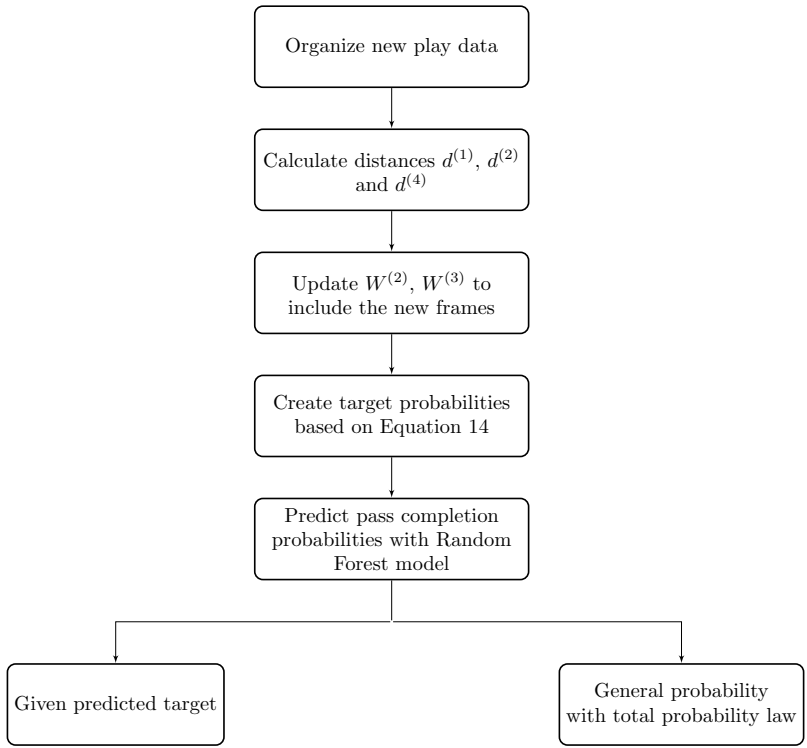}
\caption{Step-by-step algorithm of how to calculate pass completion probabilities for new observations}
\label{alg}
\end{figure}

To summarise all our work, Figure \ref{alg} shows the step-by-step algorithm of what we would have to do if we obtain a new play to calculate the probabilities of pass completion.

\section{Results}
\label{chap2}

The results presented in this Section are derived from the test subsets created by the cross-validation.

We have $4$ or $5$ offensive players that can be the target in every play, which means $4$ or $5$ observations per frame. We were interested in the probabilities $P(C \mid T=i)$ described in Equation \ref{totalp} for us to be able to calculate $P(C)$ as well.

For a practical example, in the first play of the game between the Philadelphia Eagles and the Atlanta Falcons on week 1, which was a completed pass from Matt Ryan to Julio Jones for a $10$-yard gain, we present the probabilities for the first frame after the pass started in Table \ref{tableprob}. With these probabilities we have that for this frame $P(C) = 0.579$.

\begin{table}[!htbp]
\centering
\caption{Probabilities of players being the target ${P(T=i)}$ and the completion probability given that the player is the target ${P(C \mid T=i)}$, for the first frame after the pass started of the first play of the game between the Atlanta Falcons and the Philadelphia Eagles on week 1}
\label{tableprob}
\begin{tabular}{ccc}
\hline
\textbf{Player} & $\boldsymbol{P(T=i)}$ & $\boldsymbol{P(C \mid T=i)}$ \\ \hline
Julio Jones     & 0.477          & 0.786         \\
Mohamed Sanu    & 0.101          & 0.480         \\
Devonta Freeman & 0.314          & 0.436         \\
Austin Hooper   & 0.049          & 0.190         \\
Ricky Ortiz     & 0.059          & 0.158         \\ \hline
\end{tabular}
\end{table}

Given that this play was indeed a completed pass, it is expected that for the last few frames the only player with a probability to be the target is the one that really was the target. This results in $P(C) = P(C \mid T=i)$ for $i$ being the player that was the real target, which in this play was Julio Jones. He had $P(C \mid T=i) = 0.524$ in the last frame before the event of pass completed. This is demonstrated in Table \ref{tableprob2}. We can conclude that it was not an easy pass completion, with the probability of Julio Jones catching the ball around $52\%$.

\begin{table}[!htbp]
\centering
\caption{Probabilities of players to be the target and the completion probability given that the player is the target, for the frame when it was considered a completed pass in the first play of the game between the Atlanta Falcons and the Philadelphia Eagles on week 1}
\label{tableprob2}
\begin{tabular}{ccc}
\hline
\textbf{Player} & $\boldsymbol{P(T=i)}$ & $\boldsymbol{P(C \mid T=i)}$ \\ \hline
Julio Jones     & 1.000          & 0.524         \\
Mohamed Sanu    & 0.000          & 0.252         \\
Devonta Freeman & 0.000          & 0.472         \\
Austin Hooper   & 0.000          & 0.462         \\
Ricky Ortiz     & 0.000          & 0.400         \\ \hline
\end{tabular}
\end{table}

\begin{figure}[!htbp]
    \centering
    \includegraphics[width=0.45\textwidth]{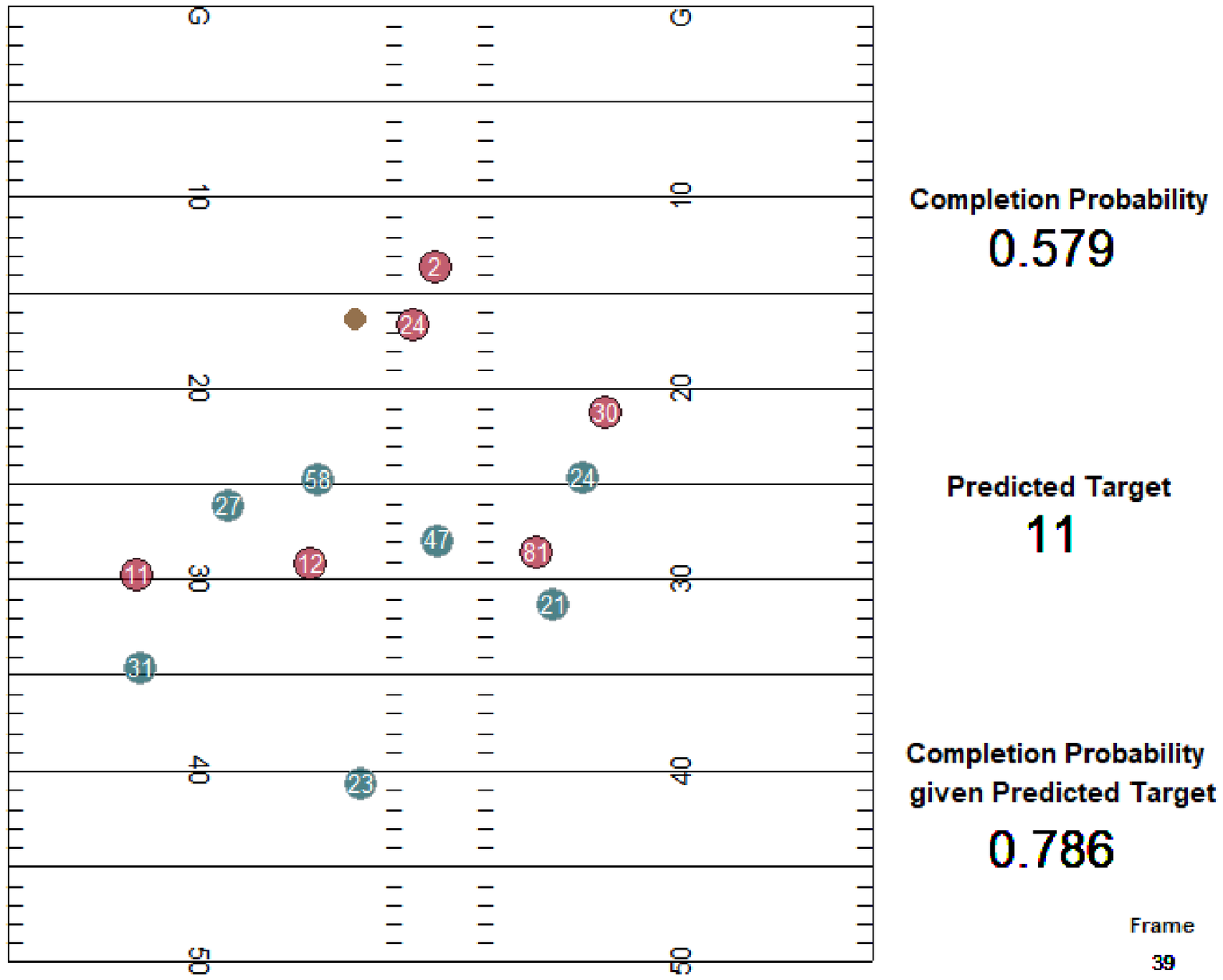}
    \includegraphics[width=0.45\textwidth]{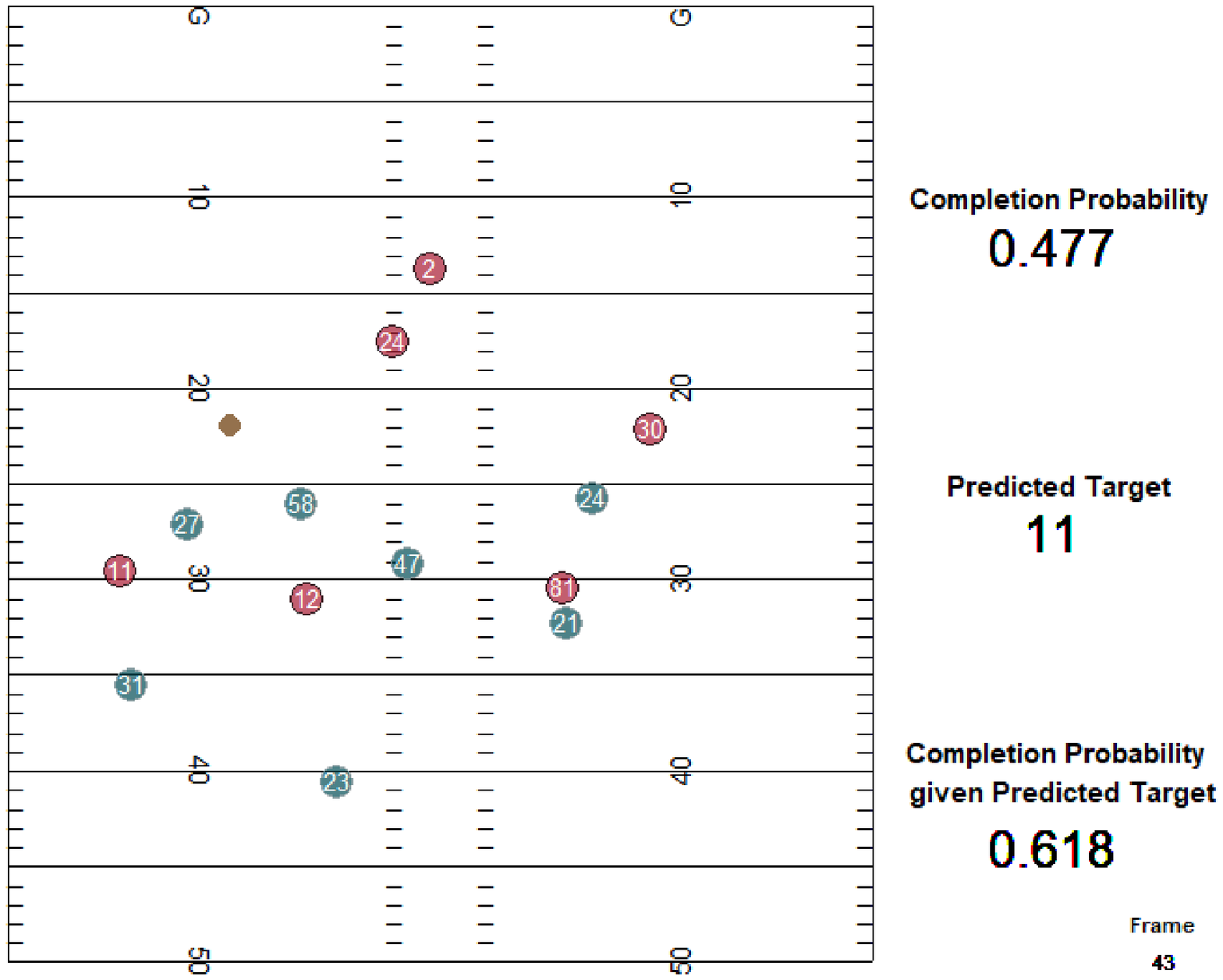}
    \includegraphics[width=0.45\textwidth]{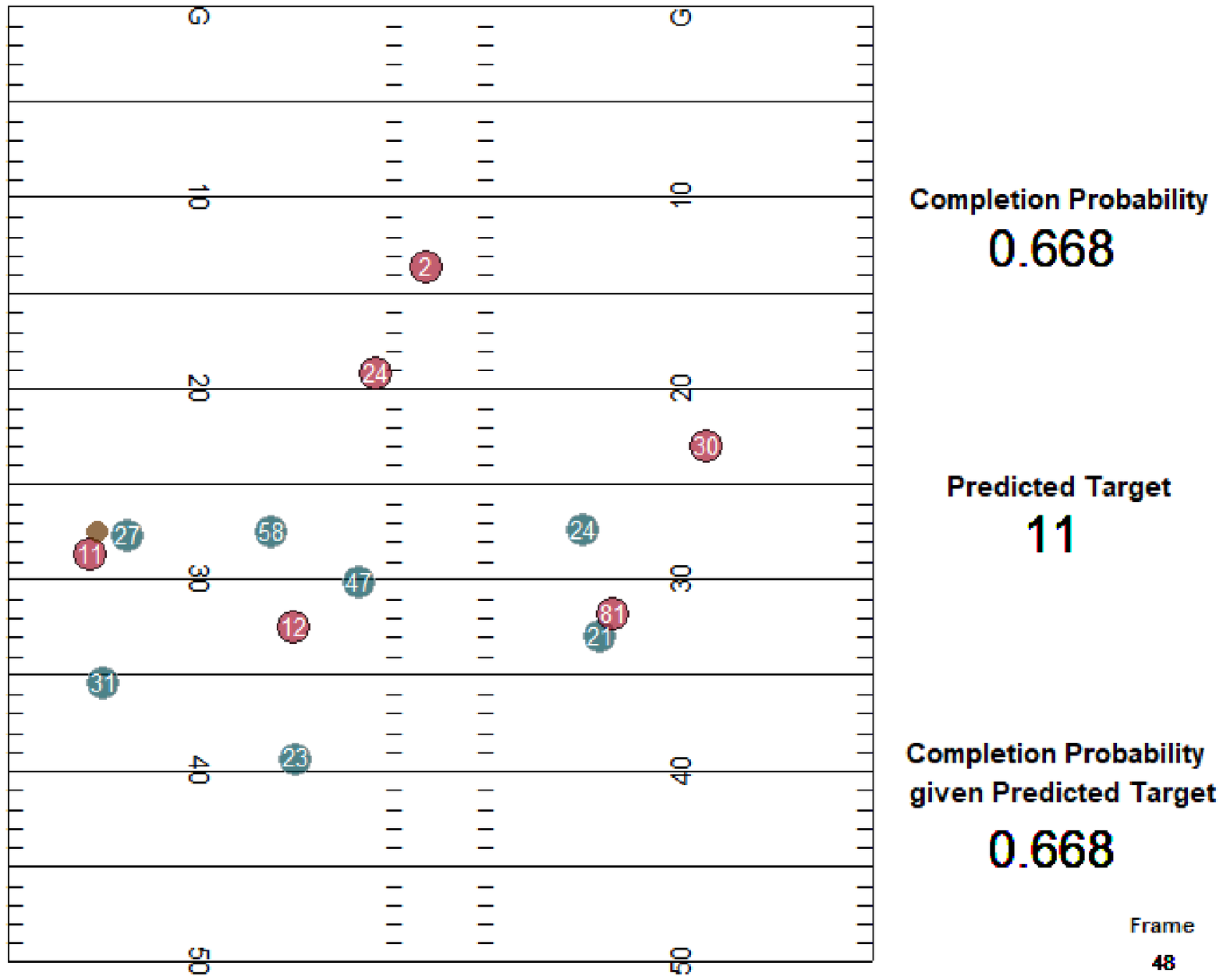}
    \includegraphics[width=0.45\textwidth]{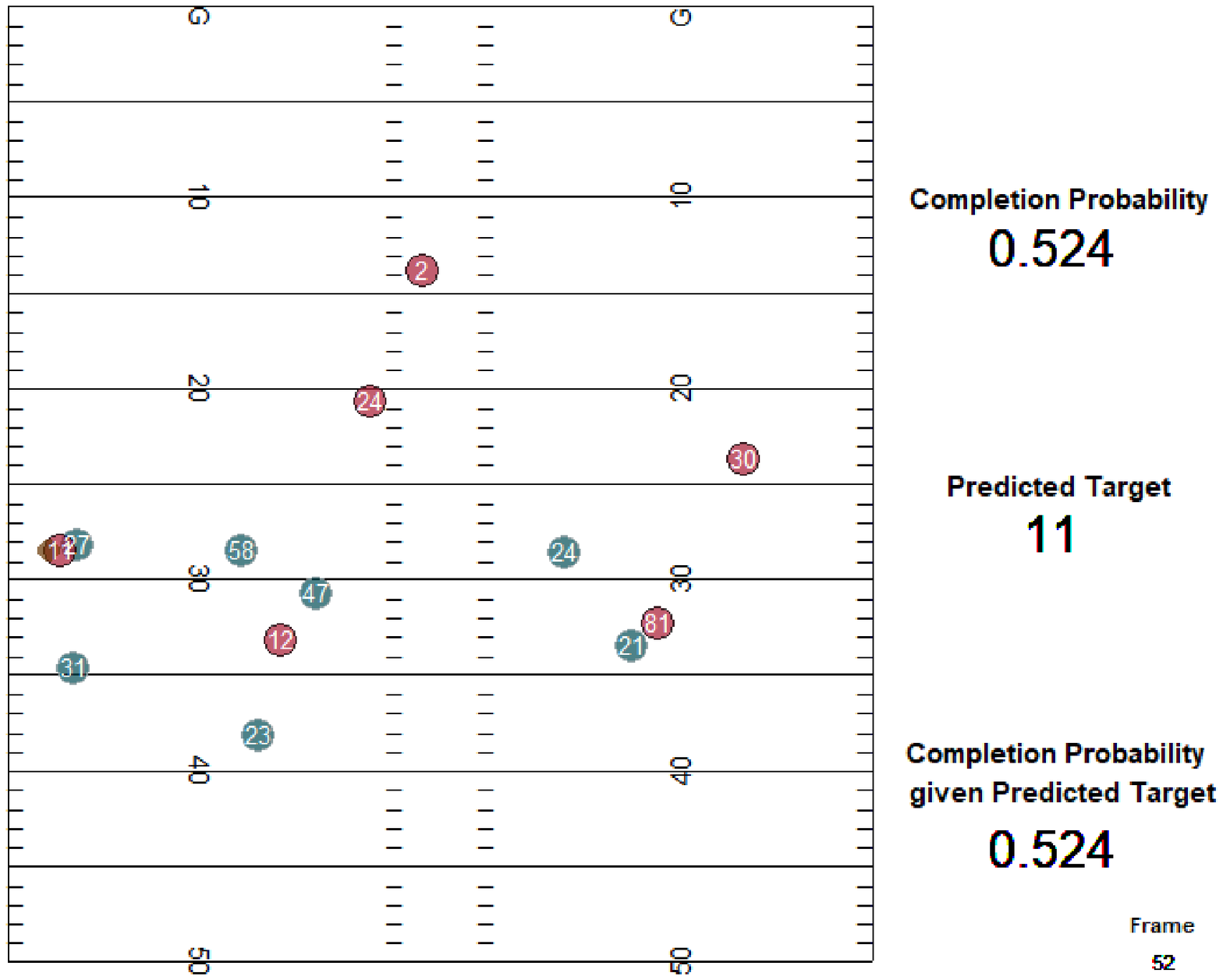}
    \caption{Frames 39, 43, 48 and 52 of a Julio Jones (\#11) reception in the first play of the game between the Atlanta Falcons and the Philadelphia Eagles on week 1 with information on the probabilities. The ball is represented in brown, the offense in red and the defense in blue.}
    \label{julio}
\end{figure}

Having the probabilities for all the frames in between, we can make animations to demonstrate the evolution of the probabilities during any play. In Figure \ref{julio} we see four frames of this Julio Jones (\#11) reception, beginning with the frame described in Table \ref{tableprob}, and ending with the one in Table \ref{tableprob2}. The information we can see on each frame are the completion probability $P(C)$, the shirt number of the player who is the predicted target (player with highest $P(T=i)$ in the frame), the completion probability given predicted target, which is the $P(C \mid T=i)$ for player $i$ with the highest $P(T=i)$ in the frame, and the number of the frame. To see the animated GIF of this whole play, visit this webpage: \url{https://media.giphy.com/media/iYoDq3oRw1JIPWLllQ/giphy.gif}.

\begin{figure}[!htbp]
    \centering
    \includegraphics[width=0.45\textwidth]{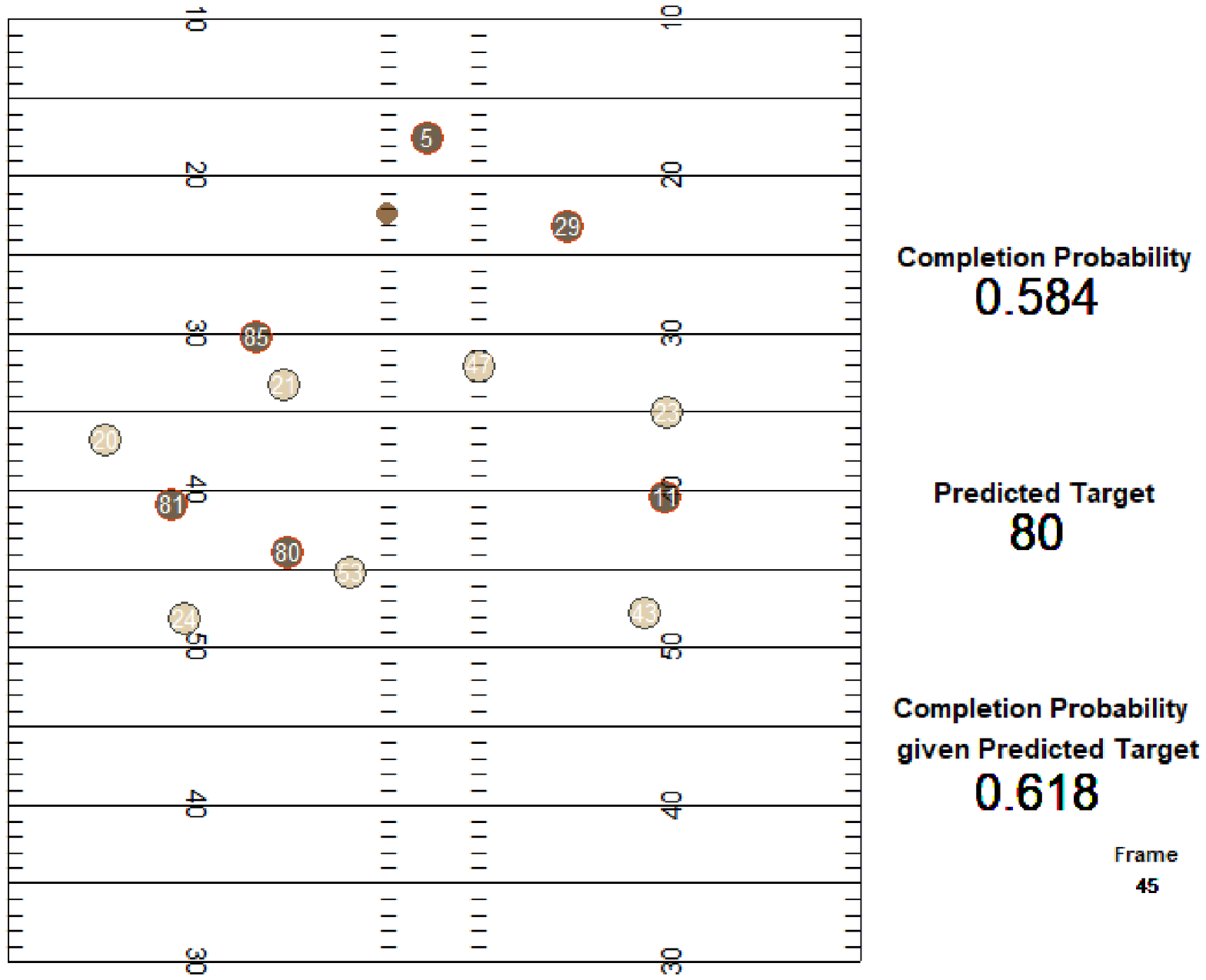}
    \includegraphics[width=0.45\textwidth]{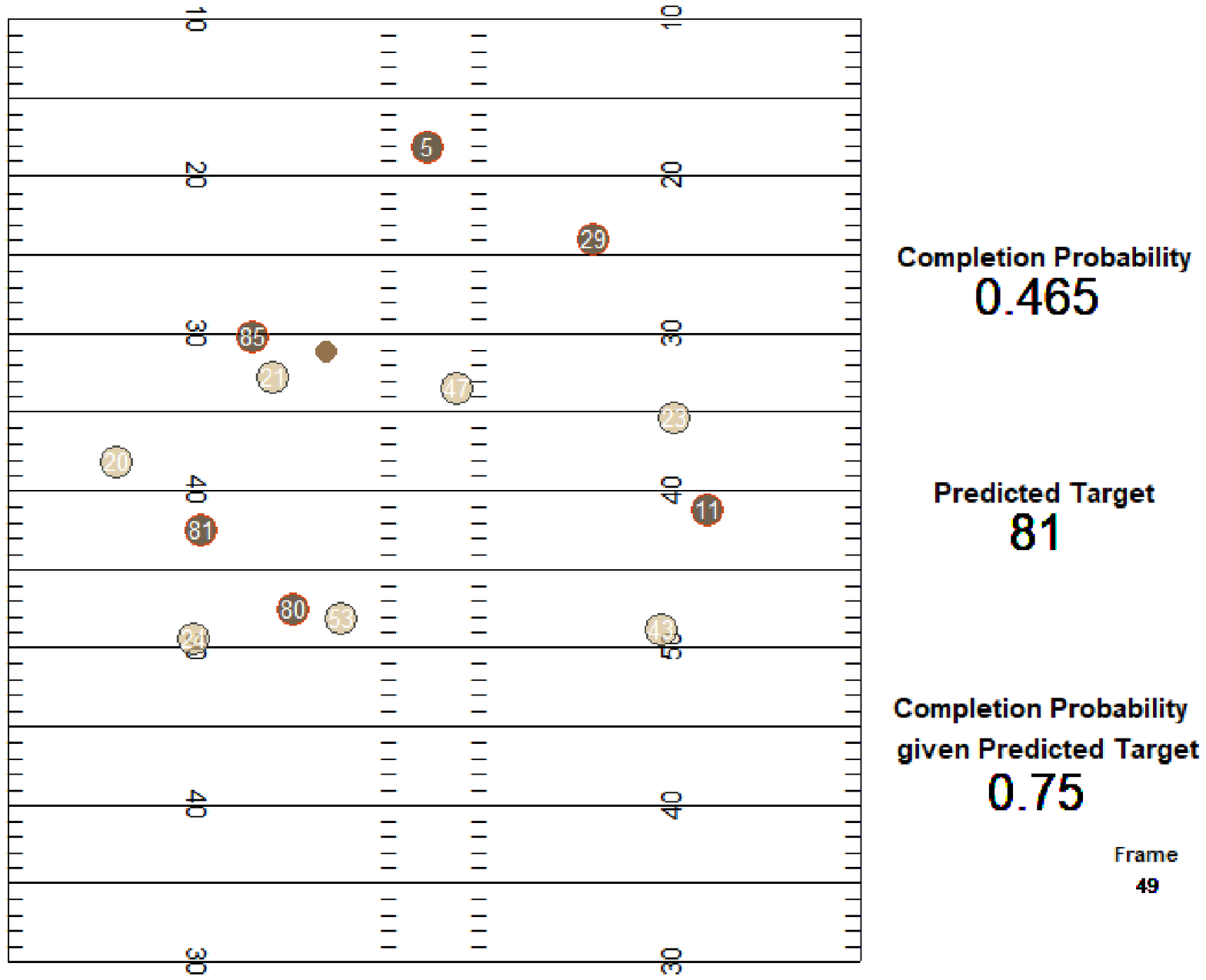}
    \includegraphics[width=0.45\textwidth]{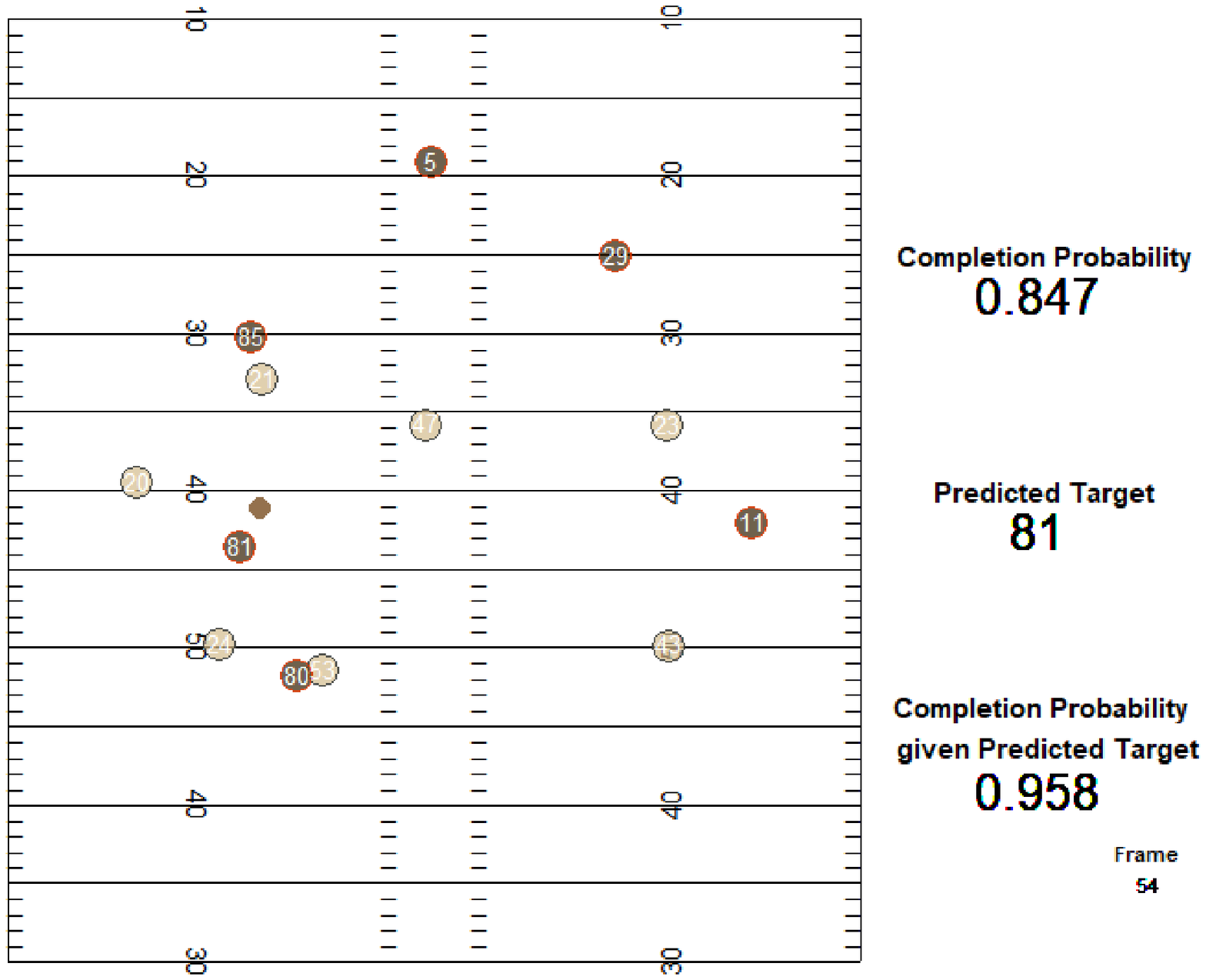}
    \includegraphics[width=0.45\textwidth]{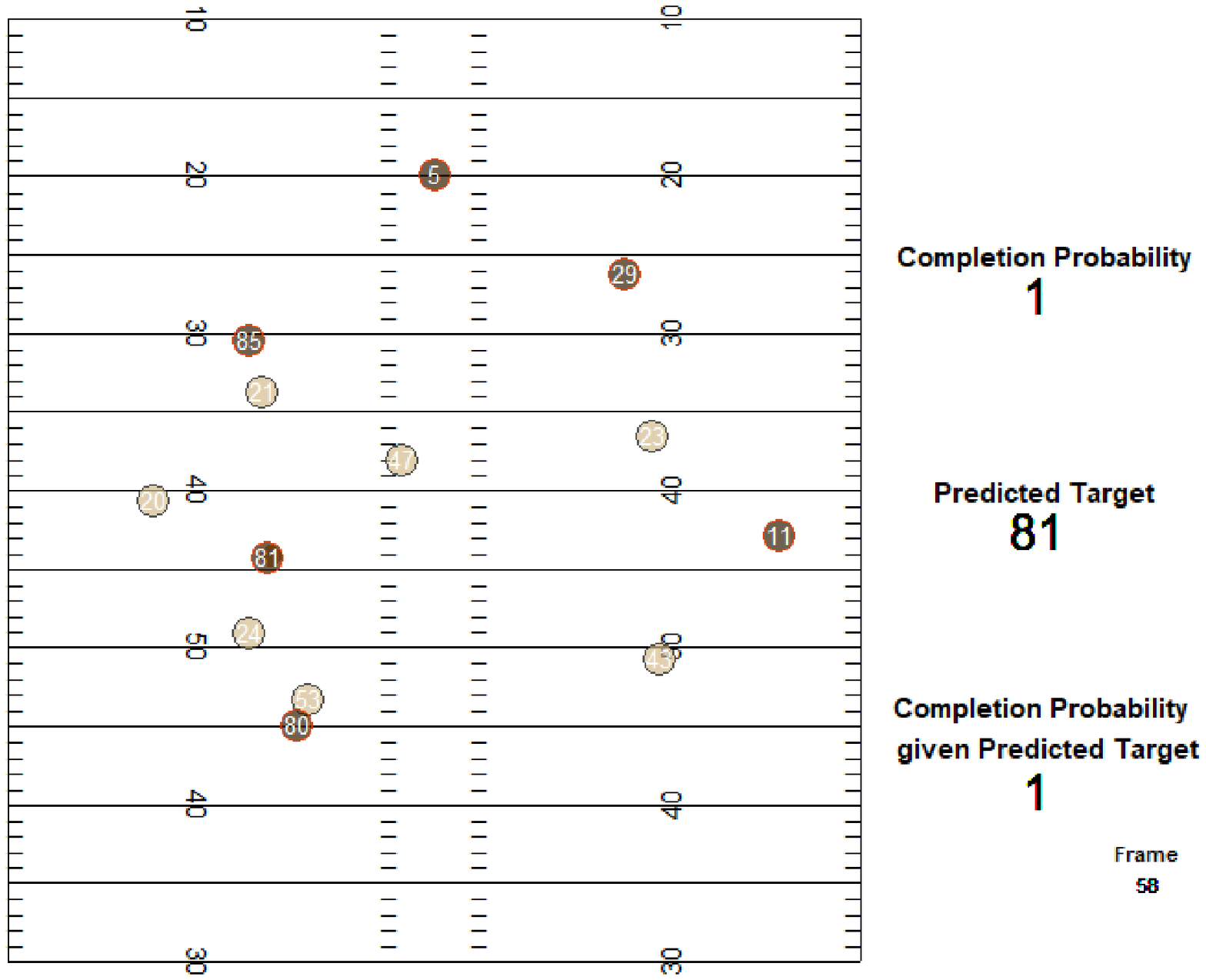}
    \caption{Frames 45, 49, 54 and 58 of a Rashard Higgins (\#81) reception in the game between the Cleveland Browns and the New Orleans Saints on week 2. The ball is represented in brown, the offense also in brown and the defense in gold.}
    \label{cle}
\end{figure}

An example of an animation of a play that was a very probable pass completion can be seen in Figure \ref{cle}. It happened on week 2 in the game between the Cleveland Browns and the New Orleans Saints, and it was a $23$ yard pass from Tyrod Taylor (\#5) to Rashard Higgins (\#81). To see the animated GIF of this whole play, visit this webpage: \url{https://media.giphy.com/media/TFRTXh6kRrlyHbH3ZB/giphy.gif}.

\begin{figure}[!htbp]
    \centering
    \includegraphics[width=0.45\textwidth]{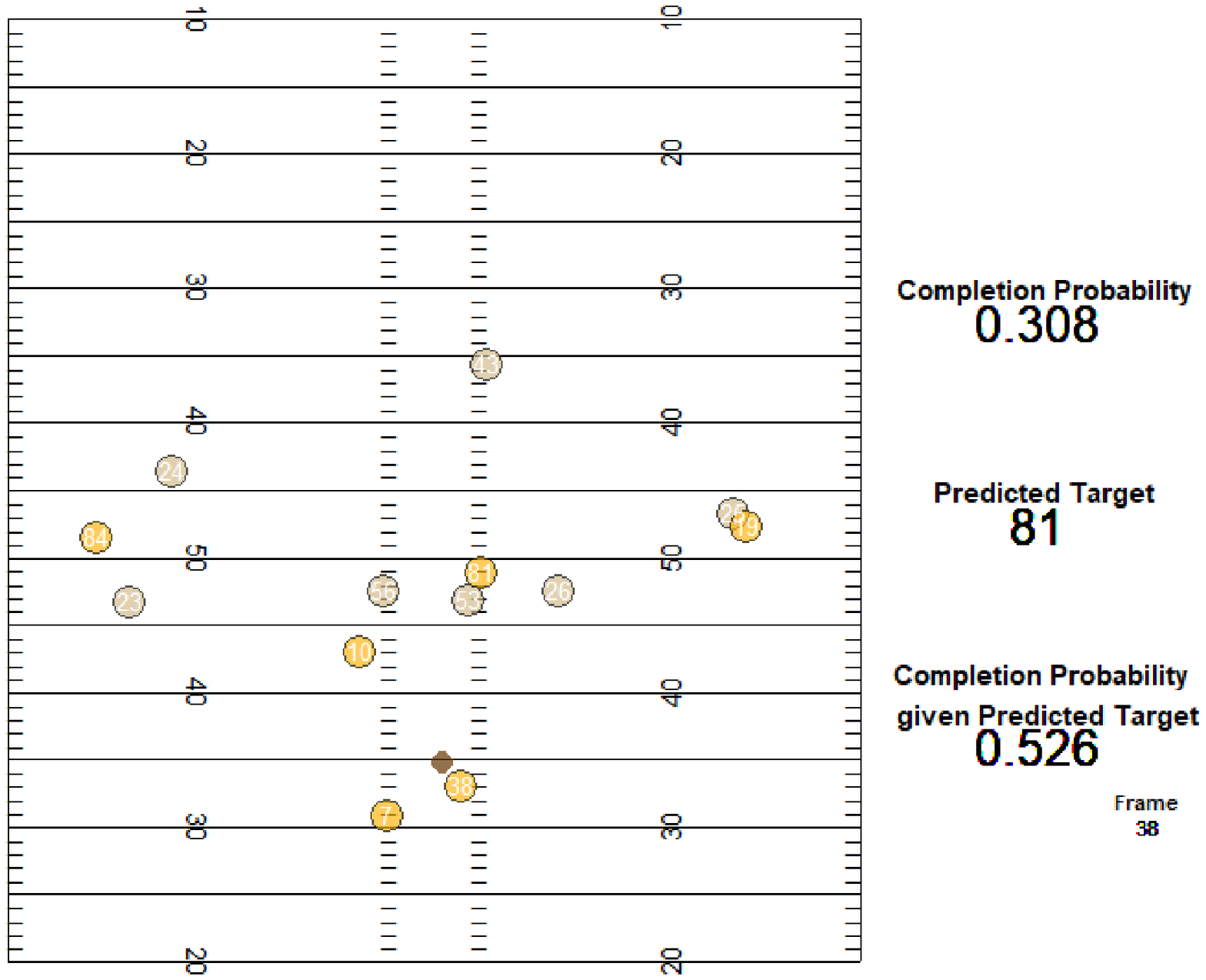}
    \includegraphics[width=0.45\textwidth]{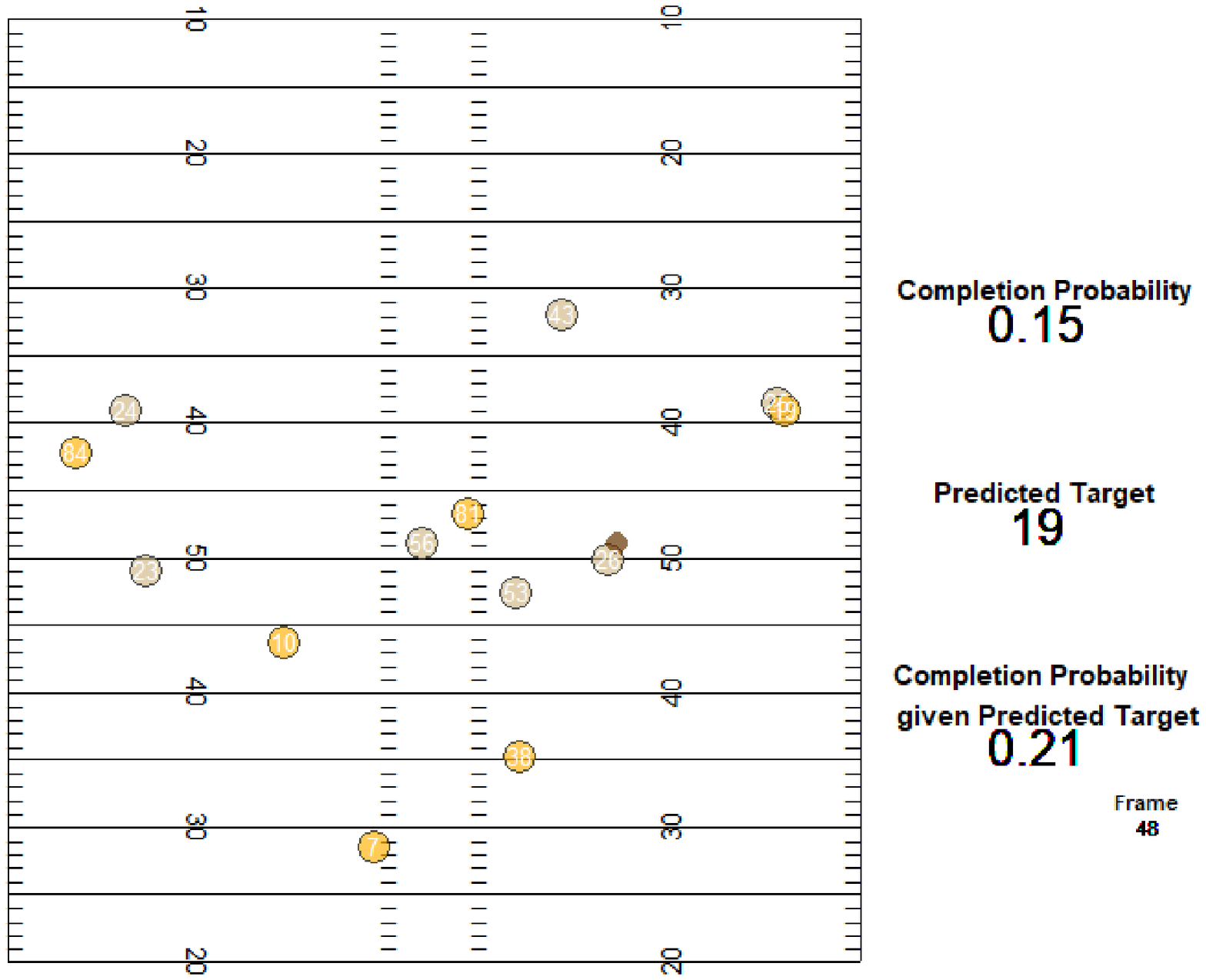}
    \includegraphics[width=0.45\textwidth]{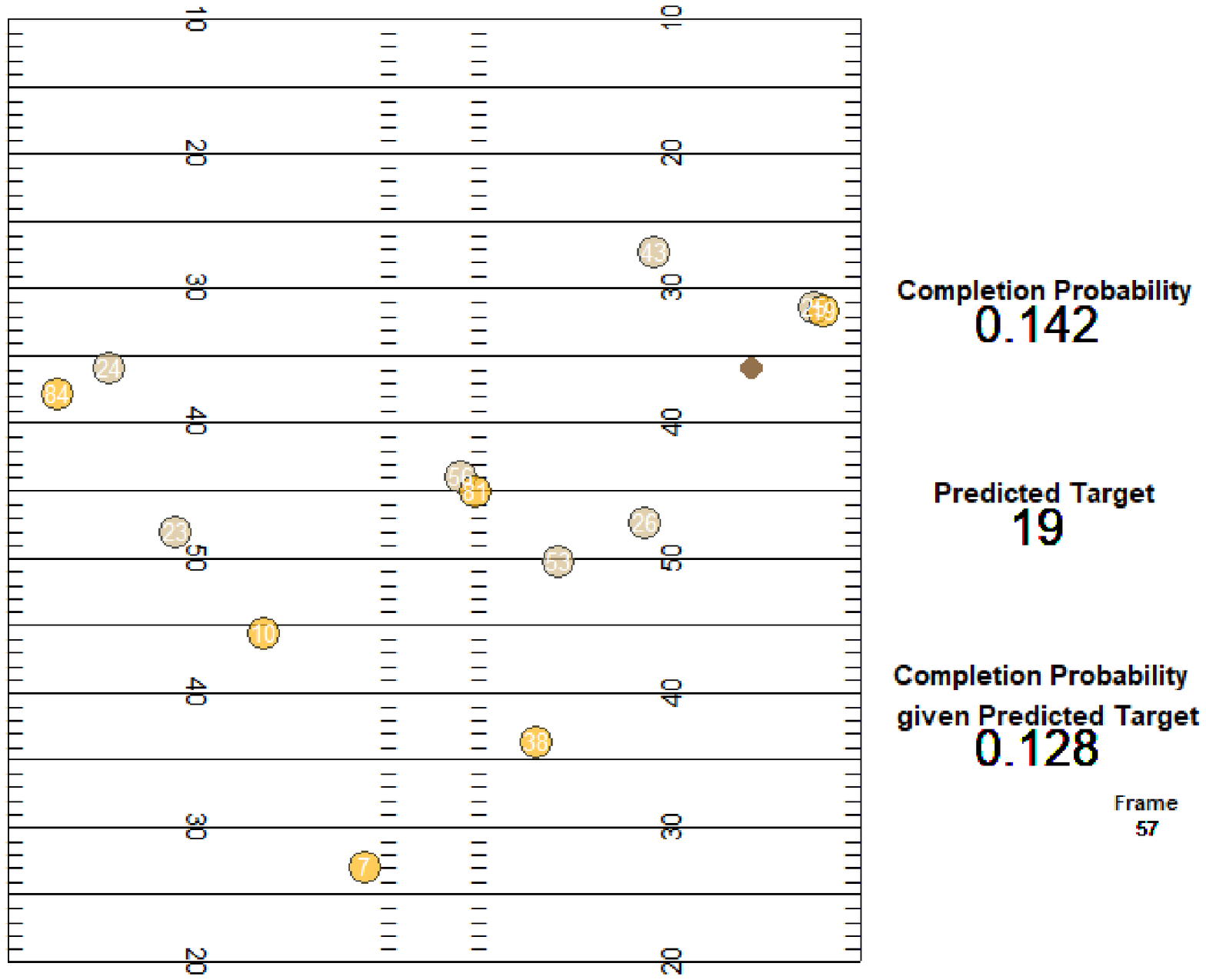}
    \includegraphics[width=0.45\textwidth]{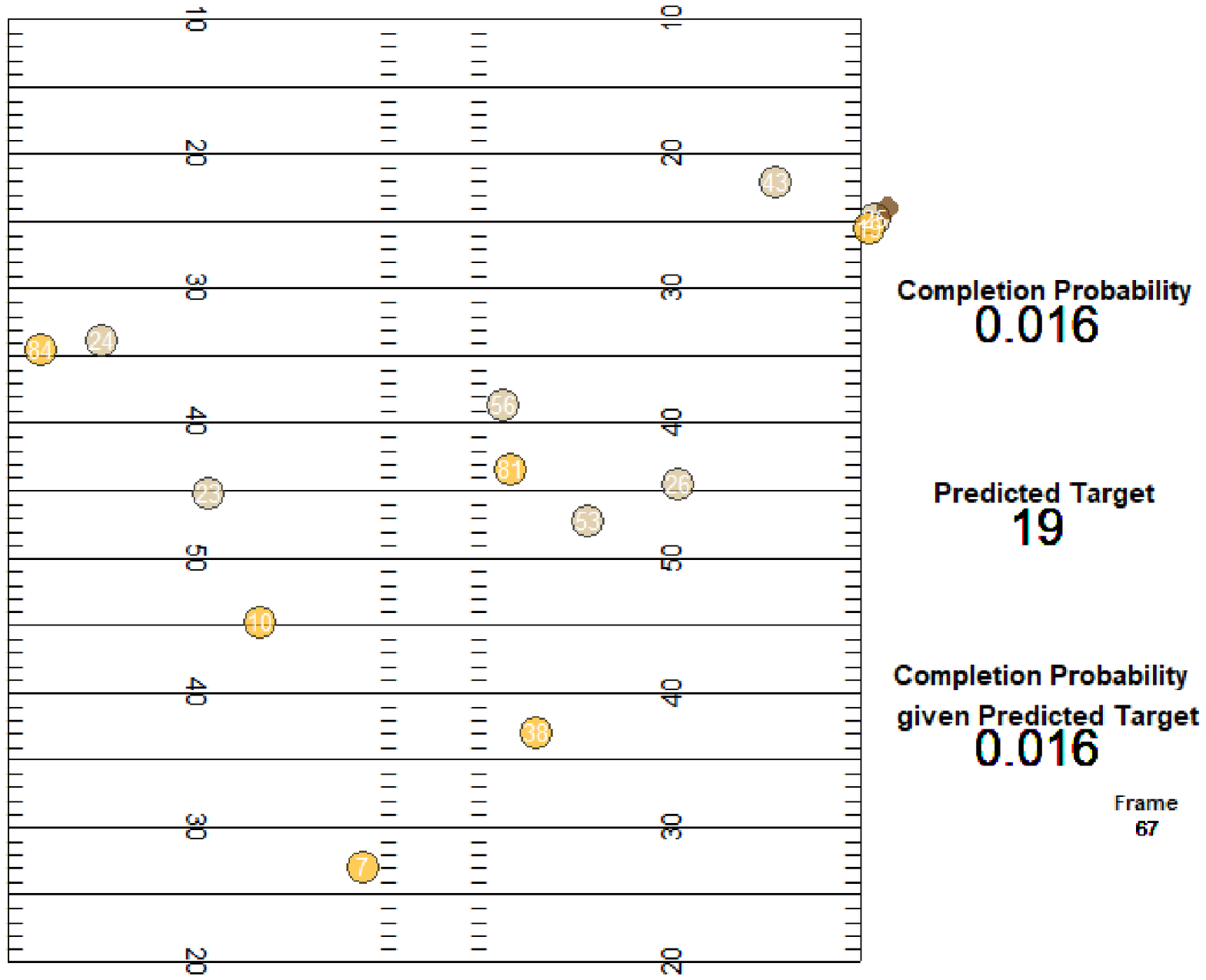}
    \caption{Frames 38, 48, 57 and 67 of an incomplete pass from Ben Roethlisberger (\#7) to JuJu Smith-Schuster (\#19) in the game between the Pittsburgh Steelers and the New Orleans Saints on week 16. The ball is represented in brown, the offense in yellow and the defense in gold.}
    \label{pit}
\end{figure}

Another example, this time of an incomplete pass that indeed had a very low completion probability can be seen in Figure \ref{pit}, where in a week 16 game between the Pittsburgh Steelers and the New Orleans Saints, Ben Roethlisberger (\#7) tried a deep pass to JuJu Smith-Schuster (\#19) but was not successful. To see the animated GIF of this whole play, visit this webpage: \url{https://media.giphy.com/media/onDmdMN2ilBzGD0X9H/giphy.gif}.

\begin{figure}[!htbp]
    \centering
    \includegraphics[width=0.45\textwidth]{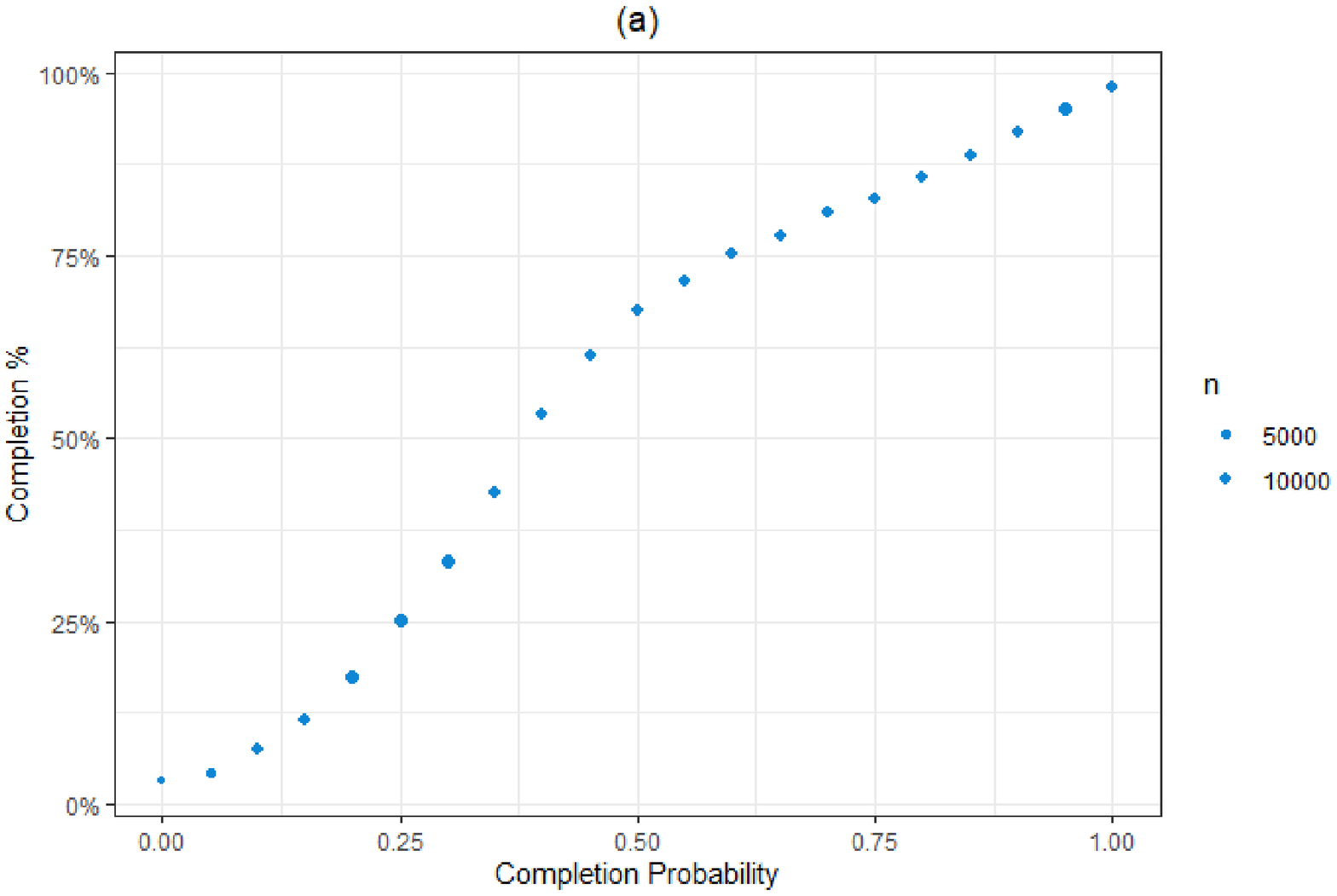}
    \includegraphics[width=0.45\textwidth]{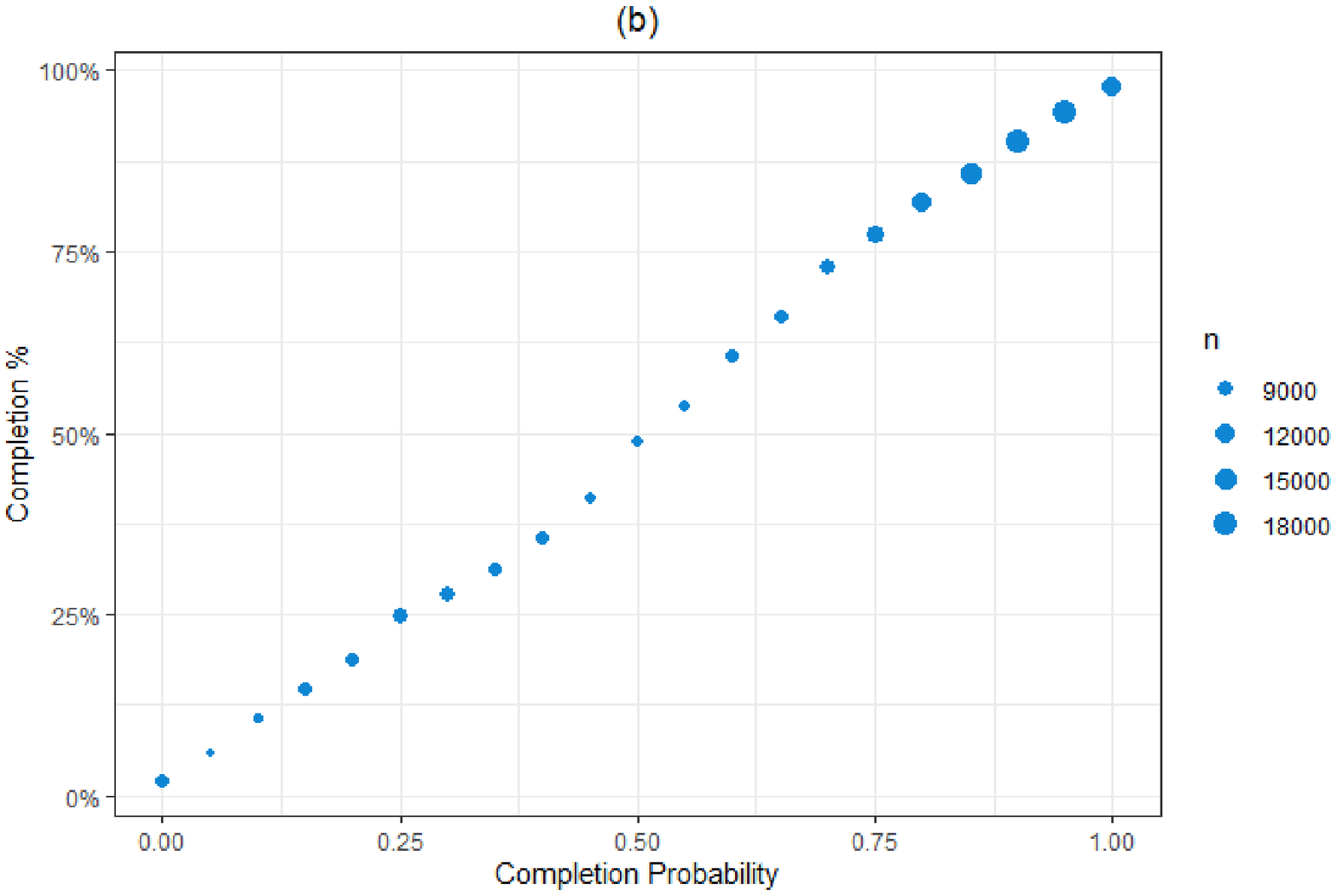}
    \includegraphics[width=0.45\textwidth]{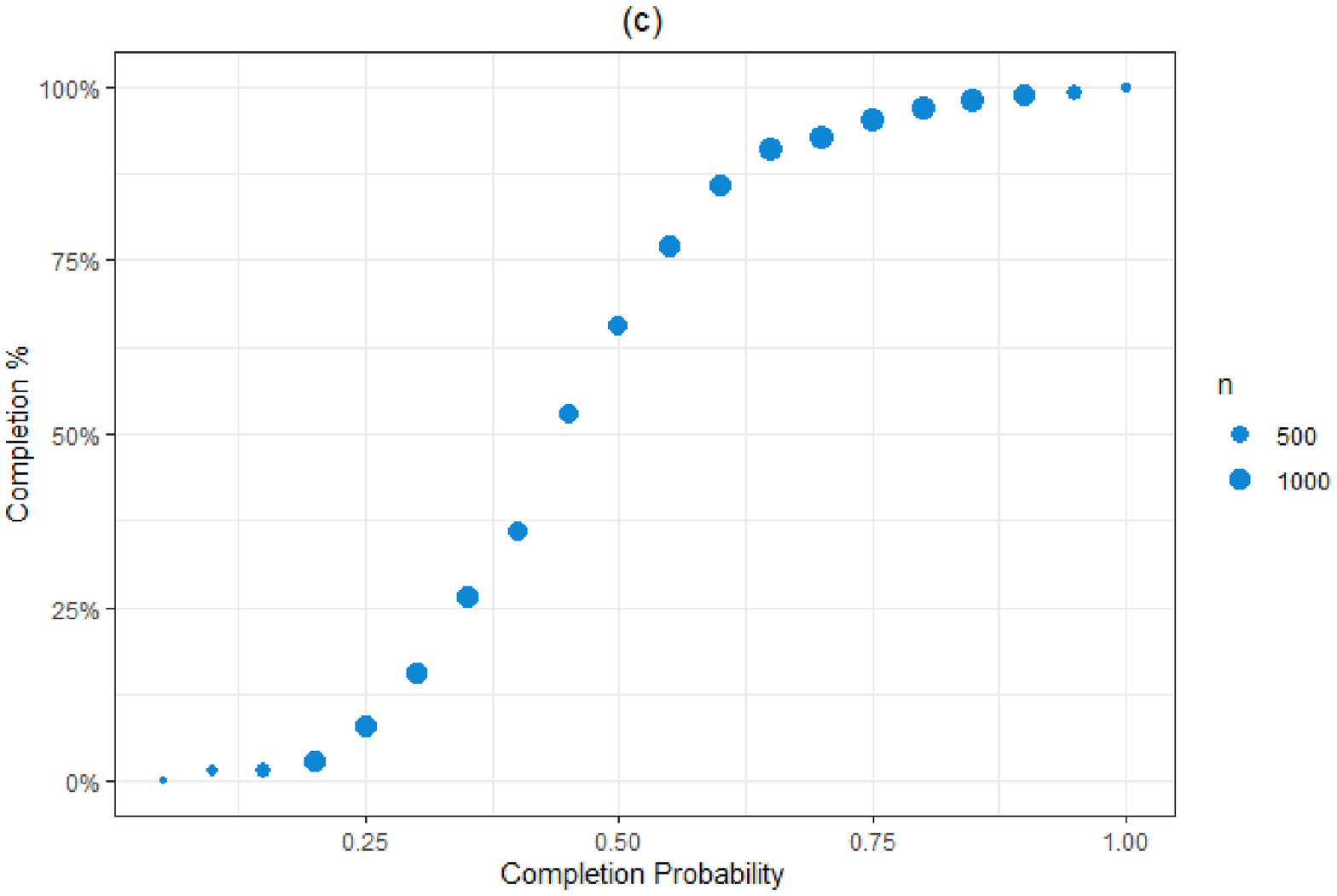}
    \includegraphics[width=0.45\textwidth]{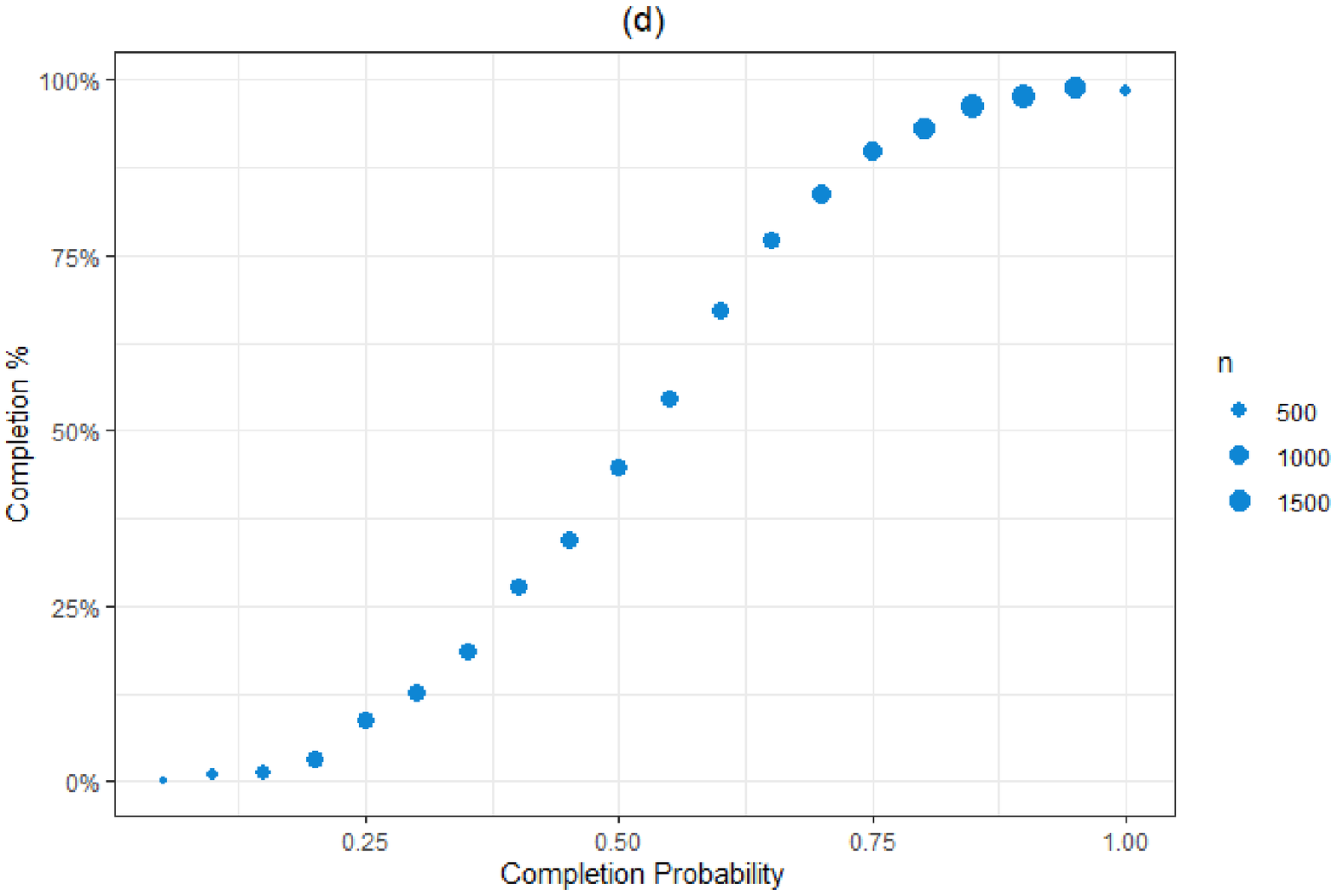}
    \caption{(a) General completion probability ($P(C)$) per frame, (b) Completion probability given predicted target ($P(C \mid T=i)$, $i$ representing the player with the highest $P(T=i)$ in the frame) per frame, (c) Average general completion probability per play and (d) Average completion probability given predicted target per play versus completion percentage}
    \label{result}
\end{figure}

In Figure \ref{result} we can see effectively how the completion probability we calculated is very relevant by plotting it versus the completion percentage. When we use the completion probability per frame, we have multiple observations for the same play, so the completion percentage represents the proportion of frames corresponding to a play that resulted in a completed pass. When we use the average completion probability, the completion percentage represents the proportion of plays that resulted in a completed pass. 

In Table \ref{conc} we show the Pearson's correlation coefficient and the Lin's concordance correlation coefficient for all the four situation shown in Figure \ref{result}. From this Table, we can see that the results are far better when comparing $P(C \mid T=i)$ to the completion percentage than when comparing $P(C)$, this shows that our probabilities to determine the player most likely to be the target are working very well. Another conclusion that we can draw is that analysing the probabilities frame by frame is better than computing an average probability of all frames on a play. This shows that our model is obtaining very accurate results even in the beginning of plays, most likely because the variables of distance to projection of the line made from the ball or the players are a very good indicator if the pass is going in the right direction and if the offensive player is well guarded by the defense or not.

\begin{table}[!htbp]
\centering
\caption{Pearson's correlation coefficient and Lin's concordance correlation coefficient for all the different setups in Figure \ref{result}}
\label{conc}
\begin{tabular}{ccc}
\hline
\textbf{Probabilities} & \textbf{Correlation} & \textbf{Concordance} \\ \hline
$P(C)$ per frame       & 0.978            & 0.958            \\ 
$P(C \mid T=i)$ per frame  & 0.998            & 0.998            \\
Average $P(C)$ per play      & 0.958            & 0.903            \\ 
Average $P(C \mid T=i)$ per play   & 0.980            & 0.942            \\ \hline
\end{tabular}
\end{table}

For the training results, if we consider a $0.5$ threshold for predictions if a pass will be completed or not for the predicted target on every frame, we get a $95.8\%$ accuracy in predicting the result of the plays.

\subsection{Next Gen Stats}

NFL player tracking, also known as Next Gen Stats, is the capture of real time location data, speed and acceleration for every player, every play on every inch of the field. Sensors throughout the stadium track tags placed on players' shoulder pads, charting individual movements within inches \cite{nextgenglossary}. The player tracking data used in this work was obtained by the NFL Next Gen Stats, and this work in general was inspired by it. During the broadcast of NFL games, many different statistics obtained by the Next Gen Stats team are shown on screen for the audience.

Documented statistics of completion probability calculated by the NFL Next Gen Stats are very hard to find. The only data we found from the 2018 season are presented in an article on the NFL website \cite{nextgen}, where they talk about the three most improbable catches of the first week (\url{https://www.nfl.com/news/next-gen-stats-introduction-to-completion-probability-0ap3000000964655}). These are the only plays we can compare the results from our work with, but unfortunately, one of these three plays is from one of three games that are missing from the database (Denver Broncos vs. Seattle Seahawks), so we actually have results from our model only for two of these plays.

\begin{figure}[!htbp]
    \centering
    \includegraphics[width=0.45\textwidth]{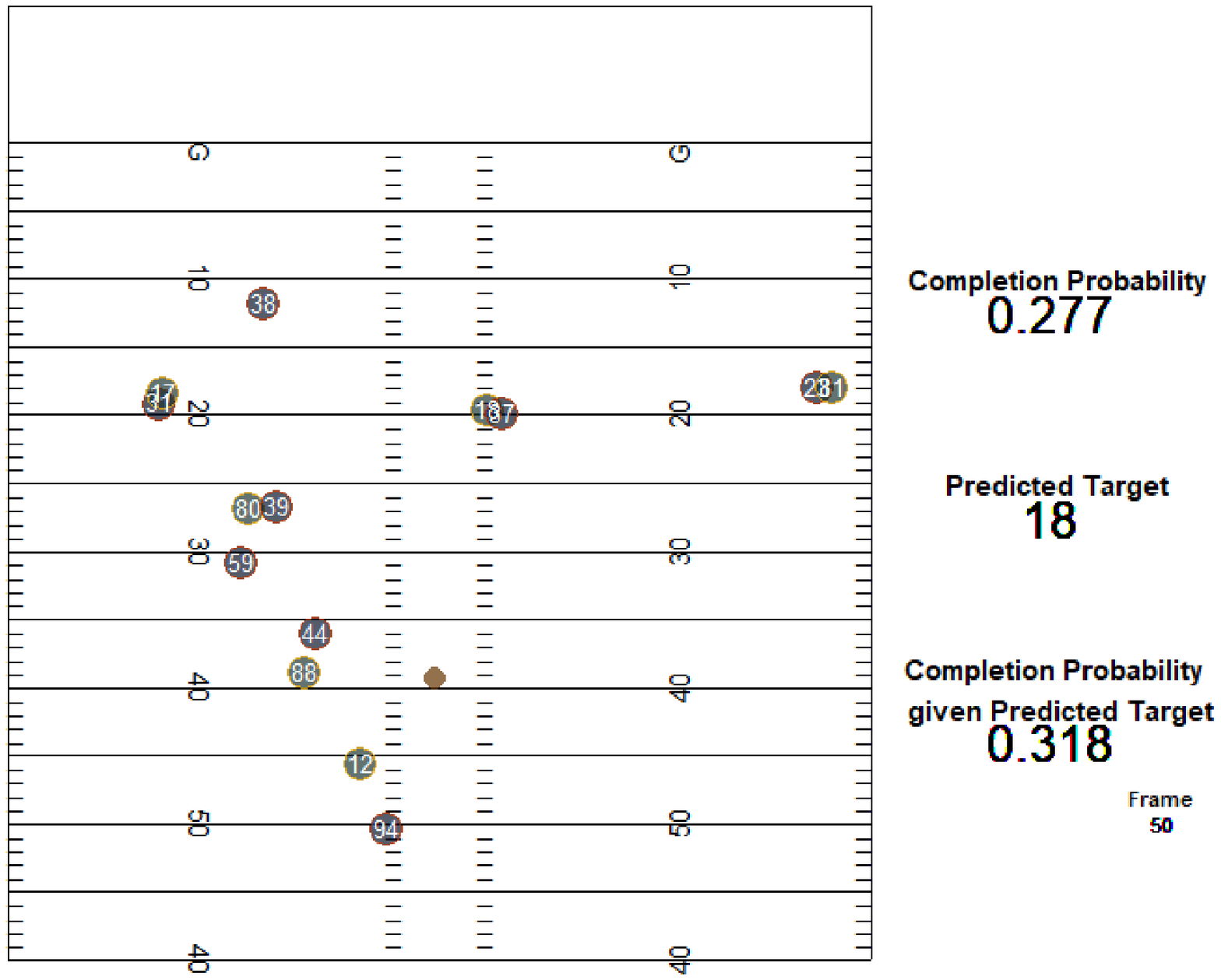}
    \includegraphics[width=0.45\textwidth]{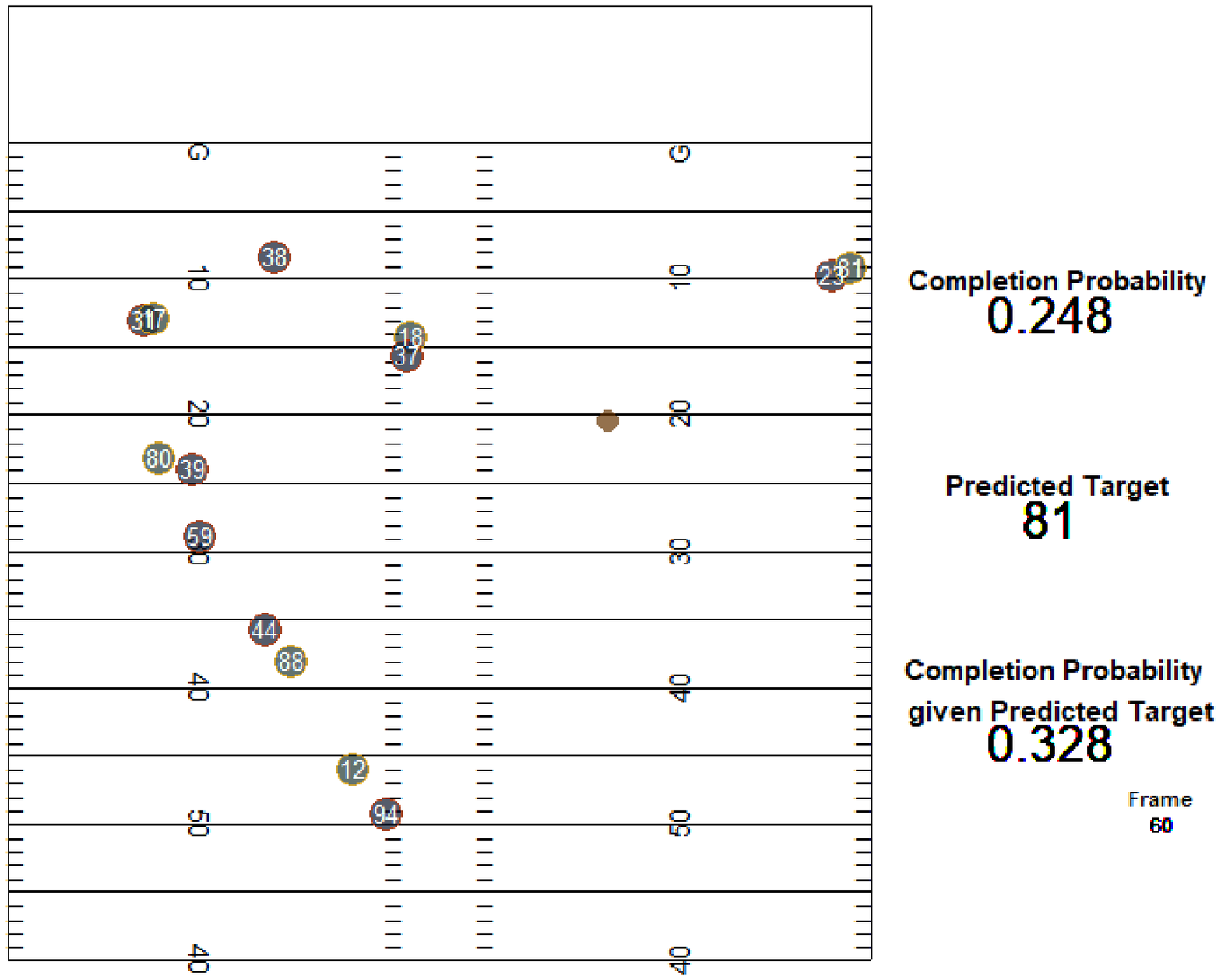}
    \includegraphics[width=0.45\textwidth]{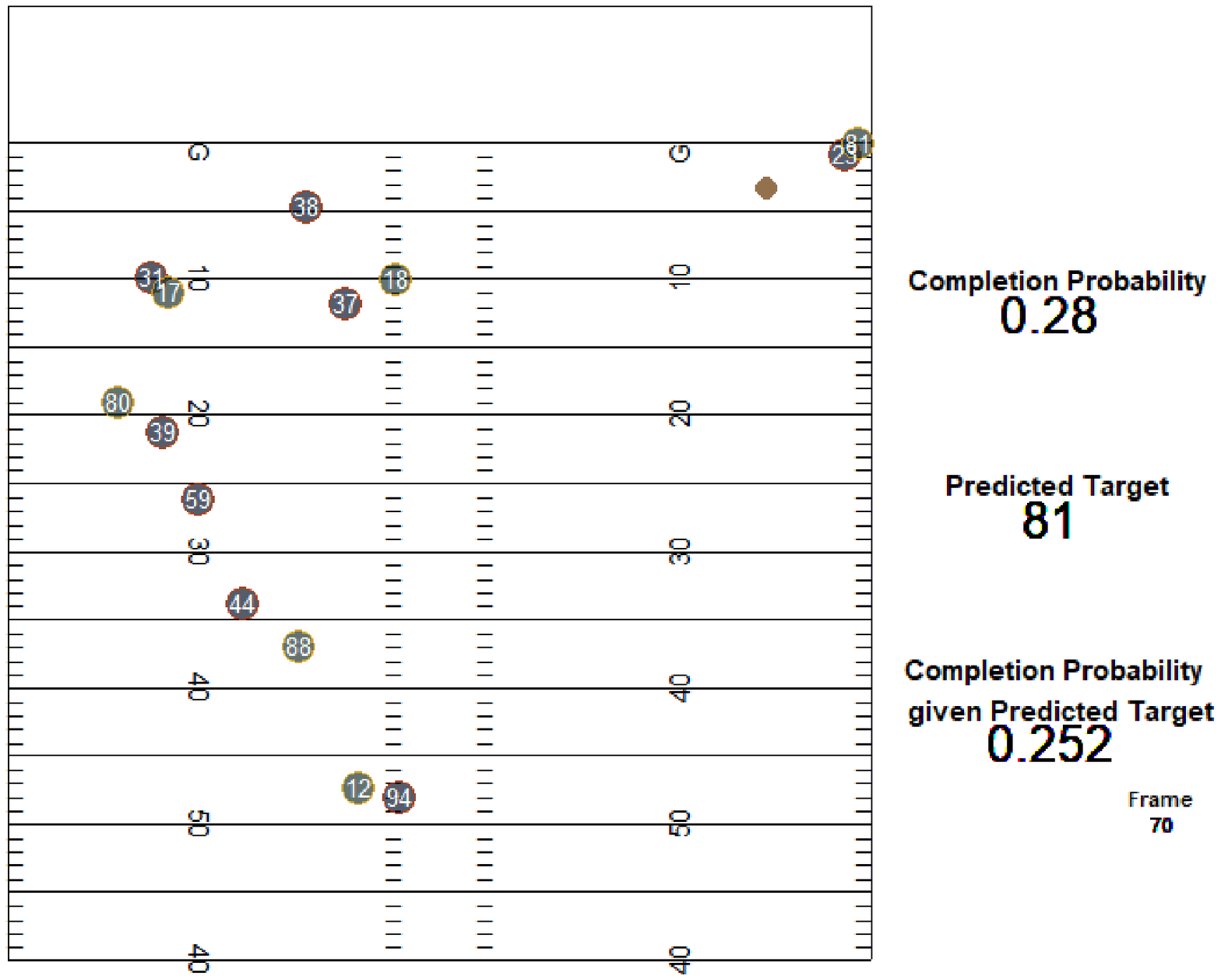}
    \includegraphics[width=0.45\textwidth]{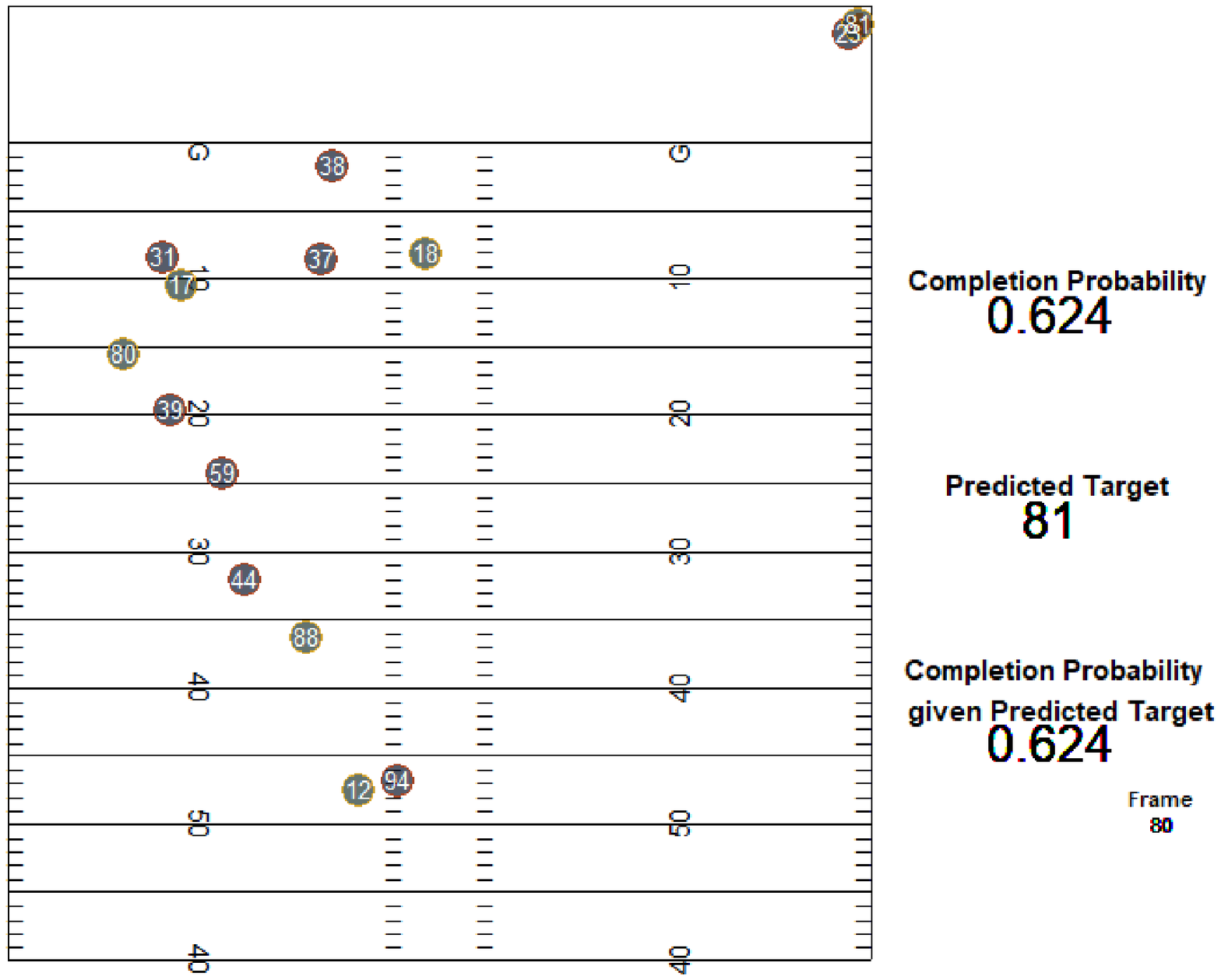}
    \caption{Frames 50, 60, 70 and 80 of the Geronimo Allison (\#81) touchdown for the Green Bay Packers versus the Chicago Bears on week 1. The ball is represented in brown, the offense in green and the defense in navy.}
    \label{geronimo}
\end{figure}

Figure \ref{geronimo} shows four frames of the first play described in the NFL article, a 39-yard touchdown pass from Aaron Rodgers (\#12) to Geronimo Allison (\#81) in the game between the Green Bay Packers and the Chicago Bears.

\begin{figure}[!htbp]
    \centering
    \includegraphics[width=0.9\textwidth]{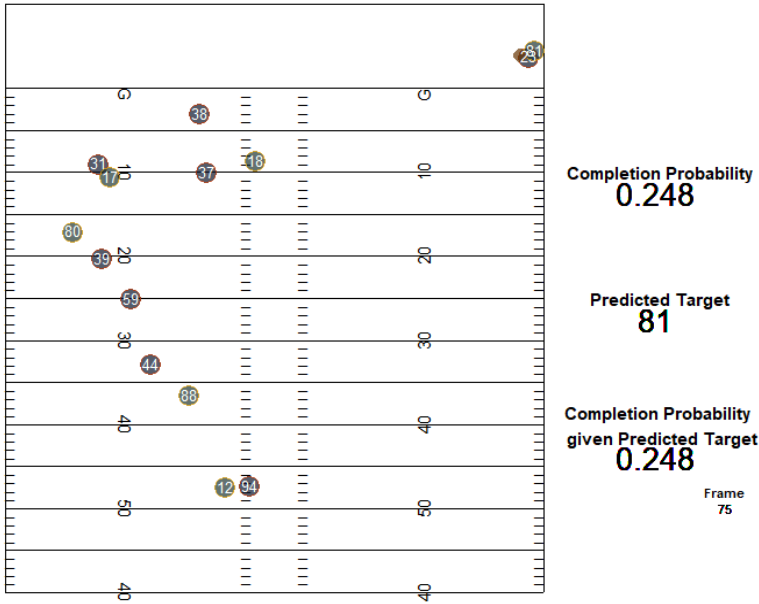}
    \caption{Frame 75 of the Geronimo Allison (\#81) touchdown for the Green Bay Packers versus the Chicago Bears on week 1. The ball is represented in brown, the offense in green and the defense in navy.}
    \label{geronimo2}
\end{figure}

In the data we used, the frame in which the play was deemed a completed pass was frame $80$, the last one shown in Figure \ref{geronimo}. From our approach, we obtained a $62.4\%$ completion probability. But if we consider a few frames earlier, such as frame number $75$, displayed in Figure \ref{geronimo2}, we see that the ball is already in range of Geronimo Allison and the completion probability is just $24.8\%$. To see the animated GIF of this entire play, visit this webpage: \url{https://media.giphy.com/media/5S0uAVk2gyVgxp072w/giphy.gif}.

\begin{figure}[!htbp]
    \centering
    \includegraphics[width=0.45\textwidth]{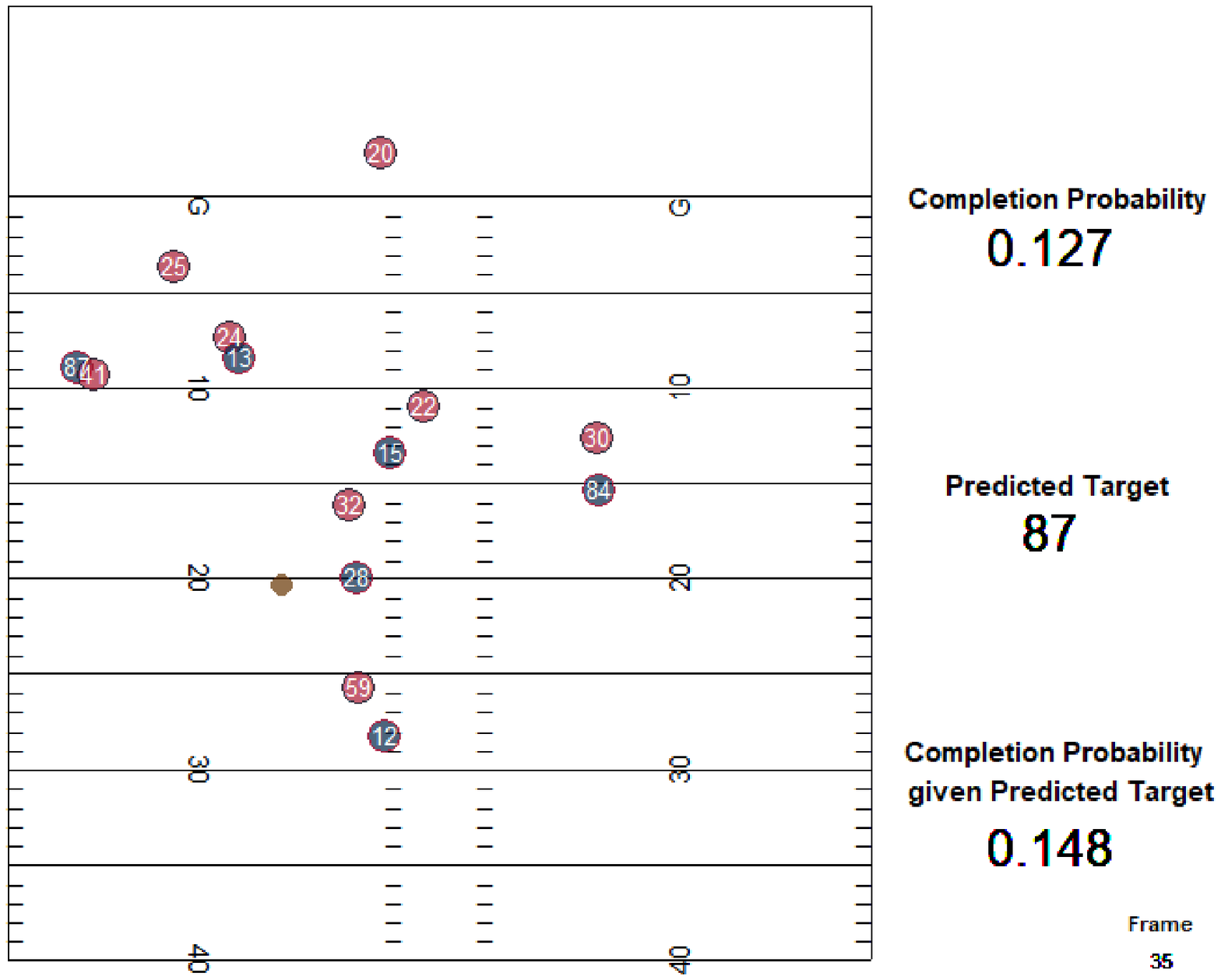}
    \includegraphics[width=0.45\textwidth]{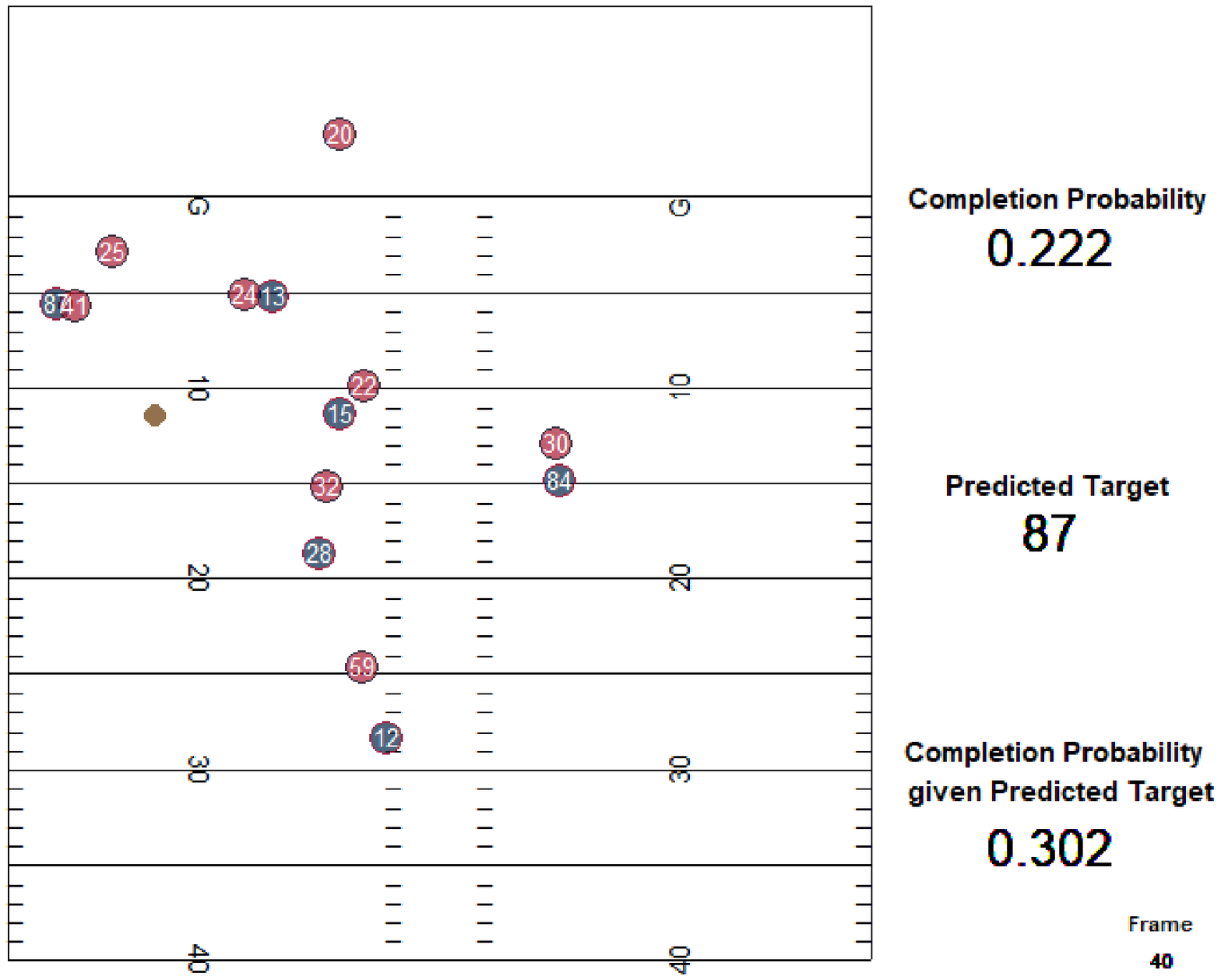}
    \includegraphics[width=0.45\textwidth]{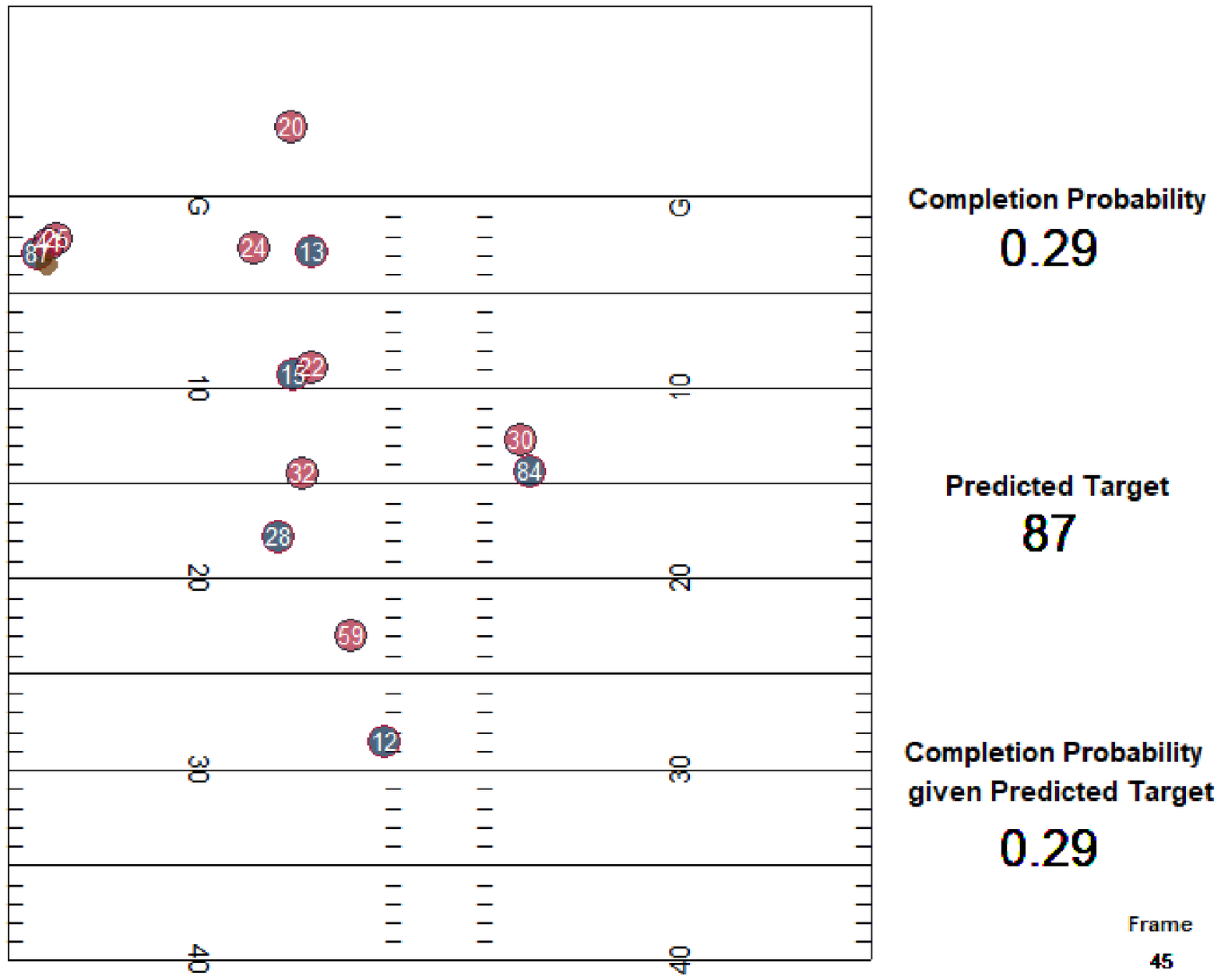}
    \includegraphics[width=0.45\textwidth]{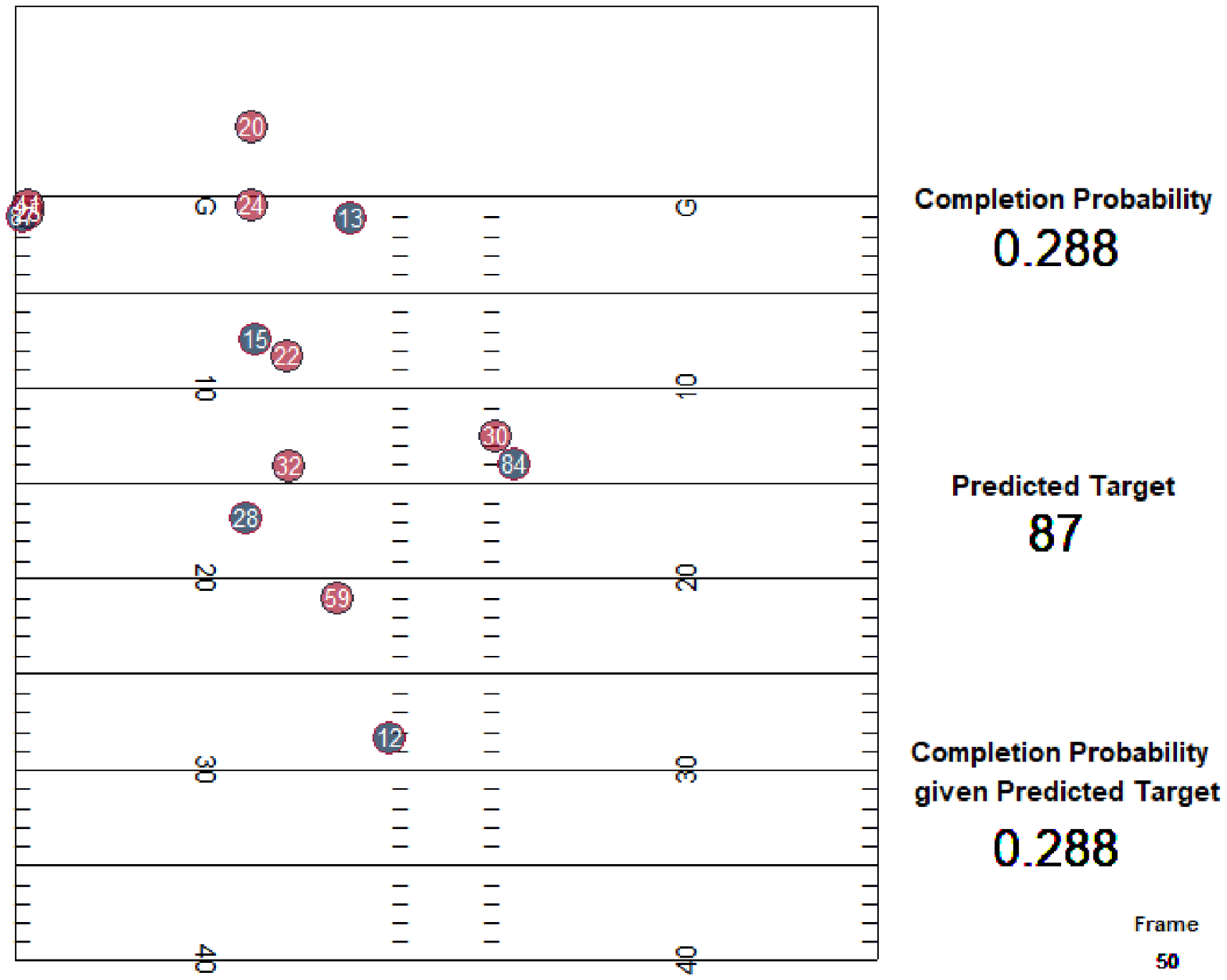}
    \caption{Frames 35, 40, 45 and 50 of the Rob Gronkowski (\#87) touchdown for the New England Patriots versus the Houston Texans on week 1. The ball is represented in brown, the offense in blue and the defense in red.}
    \label{gronk}
\end{figure}

The other play we can compare with the article is a touchdown from Tom Brady (\#12) to Rob Gronkowski (\#87) for the New England Patriots against the Houston Texans. We show four frames of this play in Figure \ref{gronk}. We can see that the final completion probability obtained from our framework was $28.8\%$. To see the animated GIF of this whole play, visit this webpage: \url{https://media.giphy.com/media/1JlGpszyaDUuHdatBS/giphy.gif}.

\begin{figure}[!htbp]
  \centering
  \includegraphics[width=0.9\textwidth]{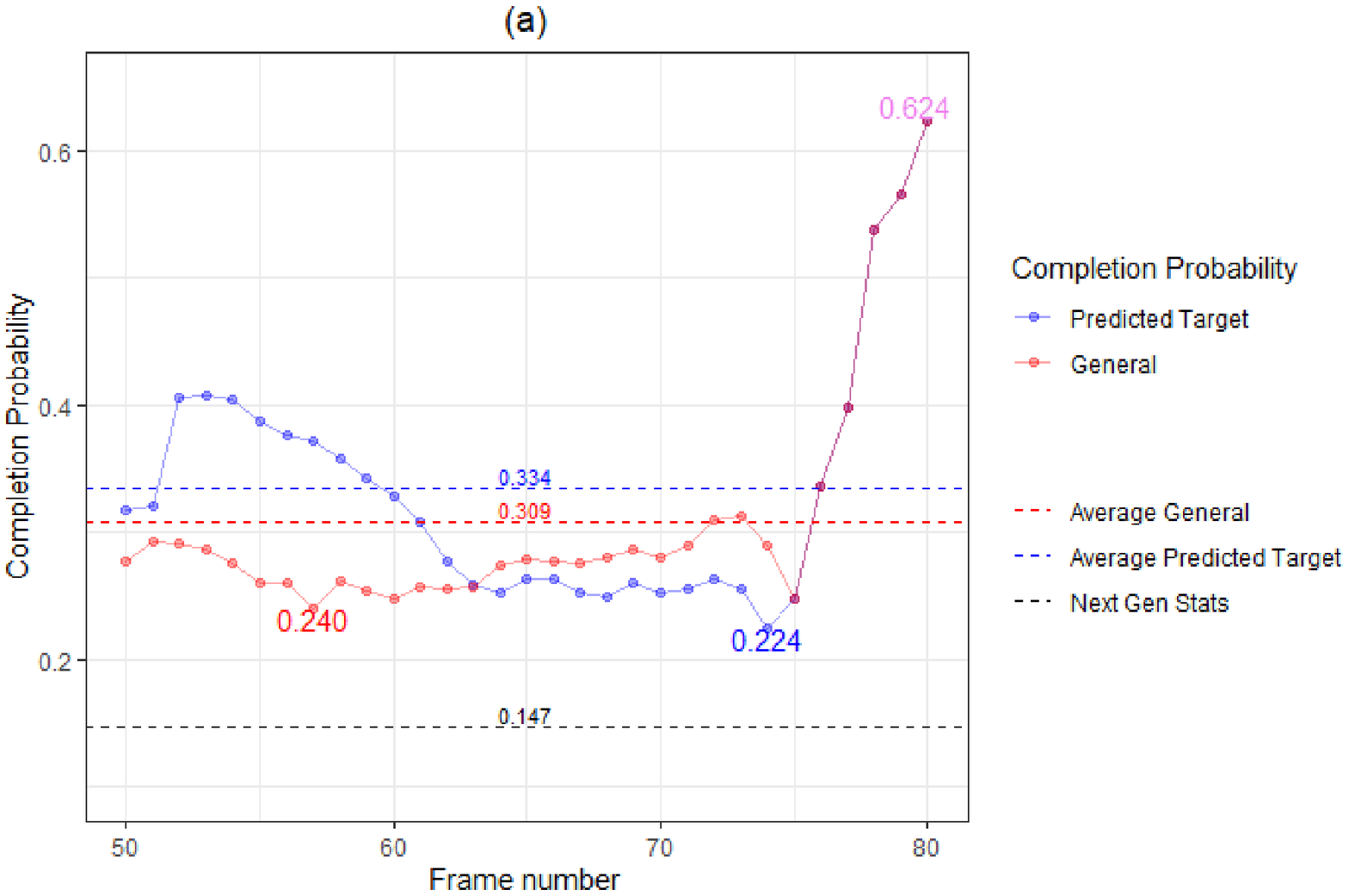}
  \includegraphics[width=0.9\textwidth]{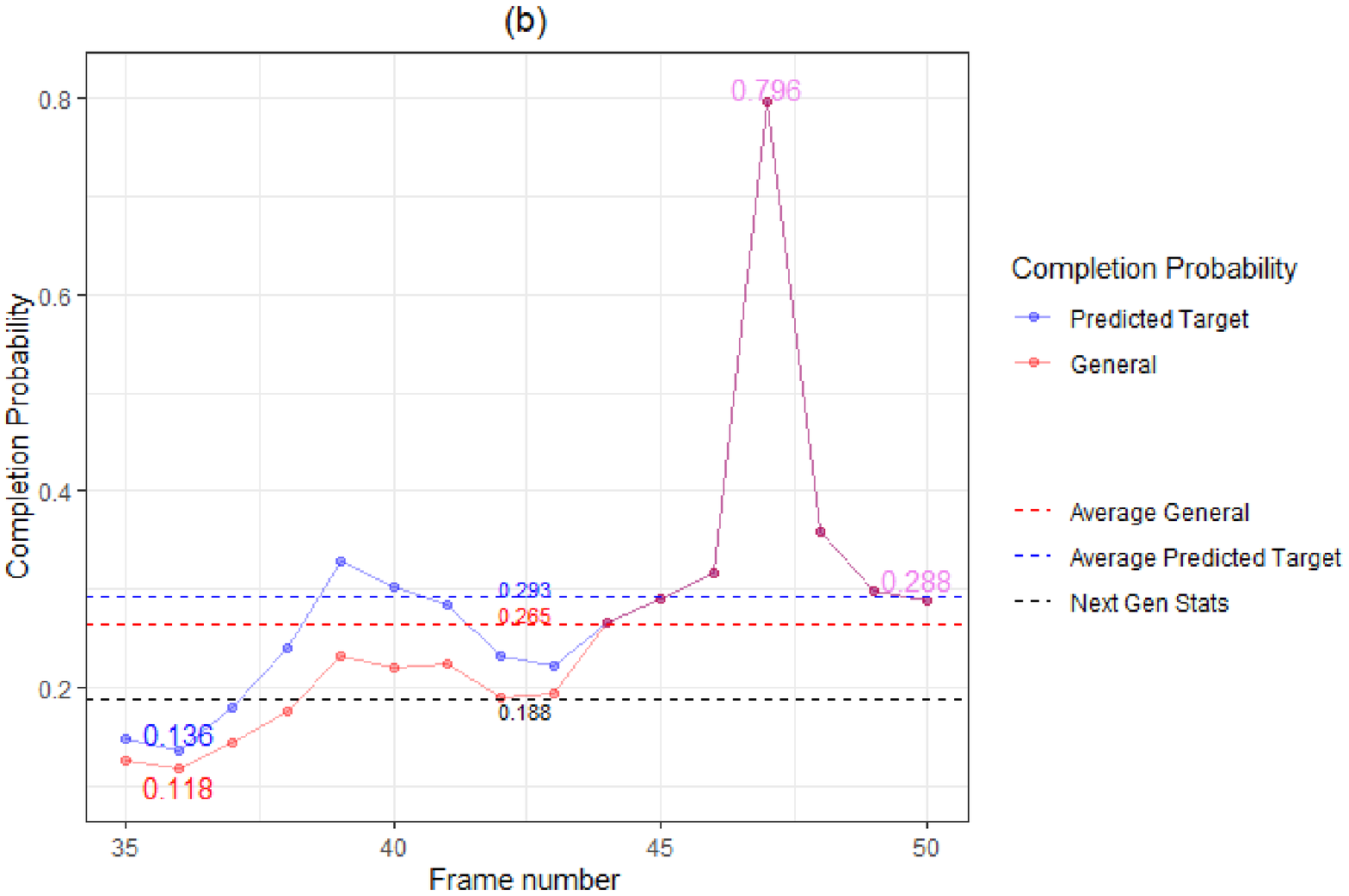}
  \caption{Evolution of completion probabilities $P(C)$ and $P(C \mid T=i)$ for all the frames during the pass in the (a) Geronimo Allison and (b) Rob Gronkowski touchdowns from week 1 games}
  \label{tds}
\end{figure}

The Next Gen Stats article does not mention at which point of the play the probabilities shown were calculated; it could be at the moment of the catch or the lowest probability reached during the play, or even an average of the whole play. Also, it could be a general probability (similar to our $P(C)$) or specific to the players cited (similar to our $P(C \mid T=i)$). To cover all these possibilities we made a plot showing the evolution of our calculated completion probabilities for all the frames during the pass (see Figure \ref{tds}), for both the Geronimo Allison and the Rob Gronkowski touchdowns.

We can conclude from Figure \ref{tds} that our framework produced higher probabilities than the NFL Next Gen Stats in almost every possible scenario, and since both plays were completed passes, it can be said that our model performed better than theirs for these particular plays.

\section{Discussion}
\label{chap3}

We calculated completion probabilities of most passing plays in the NFL 2018 season. A very relevant point is that to obtain a completion probability, we only need information on the current frame and previous ones, meaning that we do not need to know how or when the play will end. This allows us to compute the probabilities for a play in real time, given that we have enough computational power and can obtain the necessary information (such as the coordinates of the players in the field).

If the vertical coordinates of the ball were also available, this could certainly be used to improve our framework even further. This is because sometimes the ball could be very close to a player when looking at the available $x$ and $y$ coordinates, but in reality the ball is very high up in the air and going in the direction of a player further up in the field.

Our empirical probabilities of players being the target of the play proved to be very effective, even in the initial frames of the passes. The distance from players to the line projection of the ball was essential to obtain these good results.

The Random Forest model proved to be vastly superior to the other ones tested to predict the completion probabilities, even though it took a lot more time to compute. The results obtained were extremely good when comparing to the real completion percentage of the passes, as shown in Figure \ref{result}. We could not find the same data for other seasons, which unfortunately made it impossible to expand our work through more games and seasons.

Further work would include an improvement on how to determine during plays if a player still has a chance to be the target or not, and possibly utilize information not available on the data to create variables to differentiate specific players based on their historical performance on the NFL and college football.

\backmatter

\section*{Declarations}

\begin{itemize}
	\item Funding: This material is part of GPS’s Master’s dissertation, produced at the Postgraduate Programme in Statistics and Agricultural Experimentation, University of Sao Paulo, Piracicaba, Brazil, which received funding from CAPES (Coordenação de Aperfeiçoamento de Pessoal de Nível Superior) under grant no. 88887.483407/2020-00.
	\item Conflict of interest/Competing interests: The authors have no conflicts of interest to declare that are relevant to the content of this article.
	\item Ethics approval: Not applicable
	\item Consent to participate: Not applicable
	\item Consent for publication: Not applicable
	\item Availability of data and materials: Databases utilized are available at \url{https://www.kaggle.com/c/nfl-big-data-bowl-2021/data}.
	\item Code availability: All R code utilized in this paper is made available at \url{https://github.com/gustavopompeu/NFLPassCompletion}.
	\item Authors' contributions: Not applicable
\end{itemize}


\bibliography{sn-bibliography}


\end{document}